\documentclass[pdflatex,sn-mathphys]{sn-jnl}% Math and Physical Sciences Reference Style
\jyear{2021}%

\theoremstyle{thmstyleone}%
\usepackage{amsmath,physics}

\theoremstyle{thmstyletwo}%

\theoremstyle{thmstylethree}%

\begin{document}

\title[M. Tanveer et al.]{Comprehensive Review On Twin Support Vector Machines}

%%=============================================================%%
%% Prefix	-> \pfx{Dr}
%% GivenName	-> \fnm{Joergen W.}
%% Particle	-> \spfx{van der} -> surname prefix
%% FamilyName	-> \sur{Ploeg}
%% Suffix	-> \sfx{IV}
%% NatureName	-> \tanm{Poet Laureate} -> Title after name
%% Degrees	-> \dgr{MSc, PhD}
%% \author*[1,2]{\pfx{Dr} \fnm{Joergen W.} \spfx{van der} \sur{Ploeg} \sfx{IV} \tanm{Poet Laureate} 
%%                 \dgr{MSc, PhD}}\email{iauthor@gmail.com}
%%=============================================================%%

\author*[1]{M. Tanveer}\email{mtanveer@iiti.ac.in}

\author[1]{T. Rajani}\email{taanirajani@gmail.com}

\author[2]{R. Rastogi}\email{reshma.khemchandani@sau.ac.in}

\author[3]{Y.H. Shao}\email{shaoyuanhai21@163.com}

\author[1]{M. A. Ganaie}\email{phd1901141006@iiti.ac.in}

\affil[1]{Department of Mathematics, Indian Institute of Technology Indore, Simrol, Indore 453552, India}

\affil[2]{Department of Computer Science, South Asian University, New Delhi, India}

\affil[3]{School of Management, Hainan University, Haikou, China, 570228}

%%==================================%%
%% sample for unstructured abstract %%
%%==================================%%

\abstract{Twin support vector machine (TWSVM) and twin support vector regression (TSVR) are newly emerging efficient machine learning techniques which offer promising solutions for classification and regression challenges respectively. TWSVM is based upon the idea to identify two nonparallel hyperplanes which classify the data points to their respective classes. It requires to solve two small sized quadratic programming problems (QPPs) in lieu of solving single large size QPP in support vector machine (SVM) while TSVR is formulated on the lines of TWSVM and requires to solve two SVM kind problems. Although there has been good research progress on these techniques; there is limited literature on the comparison of different variants of TSVR. Thus, this review presents a rigorous analysis of recent research in TWSVM and TSVR simultaneously mentioning their limitations and advantages. To begin with, we first introduce the basic theory of support vector machine, TWSVM and then focus on the various improvements and applications of TWSVM, and then we introduce TSVR and its various enhancements. Finally, we suggest future research and development prospects.}


%%\pacs[JEL Classification]{D8, H51}

%%\pacs[MSC Classification]{35A01, 65L10, 65L12, 65L20, 65L70}

\maketitle

\section{Introduction}
\label{intro}
SVM \cite{cortes1995support} is a prominent classification technique widely used since its inception. It was first introduced by Cortes and Vapnik \cite{cortes1995support} in 1995 for binary classification problems. SVM seeks to find decision hyperplanes that determine the decision boundary which can classify the data points into two classes. These decision planes are called support hyperplanes and the distance between them is maximized by solving a quadratic programming problem (QPP). SVM is computationally powerful even in non-linearly separable cases by using kernel trick \cite{cristianini2000introduction}. SVM has remarkable advantages as it utilizes the idea of structural risk minimization (SRM) principle which provides better generalization as well as reduces error in the training phase. As a result of its superior performance even in non-linear classification problems, it has been implemented in a diverse spectrum of research fields, ranging from text classification, face recognition, financial application, brain-computer interface, bio-medicine, human action recognition, horse race odds prediction and multiple instance learning \cite{tong2001support,agarwal2014prediction,scholkopf2004support, tay2001application,gupta2019financial, noble2004support, osuna1997training, hua2001support, byvatov2003support, morra2010comparison, tanveerasparse, vapnik2013nature, edelman2007adapting,poursaeidi2014robust}. SVM has also been used in feature selection methods \cite{le2017dca}. Robust SVM penalized all the samples via a new loss function for better generalization in presence of noise \cite{wang2010adjusted}. Although SVM has outperformed most other systems, it still has many limitations in dealing with complex data due to its high computational cost of solving QPPs and its performance highly depends upon the choice of kernel functions and its parameters. Many improvements have been made in the last decade to enhance the accuracy of SVM \cite{li2020generalized, li2019single}. One such critical enhancement was generalized eigenvalue proximal SVM (GEPSVM) \cite{mangasarian2001proximal,khemchandani2017generalized,shao2013improved} that led the foundation of TWSVM \cite{khemchandani2007twin,TWSVMbook}. 
The idea behind GEPSVM is to determine two nonparallel hyperplanes via solving two generalized eigenvalue problems. TWSVM \cite{khemchandani2007twin,TWSVMbook} enhanced the generalization ability of GEPSVM. TWSVM also determines two nonparallel hyperplanes and needs to solve a pair of small QPPs in lieu of solving one complex QPP in SVM. Computational cost of TWSVM is approximately one-fourth of the SVM. TWSVM is faster and has better generalization performance than SVM and GEPSVM. Authors also extended TWSVM for nonlinear classification problems by introducing kernel. But, in nonlinear cases when a kernel is used, the formulation requires to solve inverse of matrices. Also, its performance highly depends upon the selection of kernel functions and its parameters for nonlinear classification. Although TWSVM is still in its rudimentary stage, many improvements and variants have been proposed by researchers due to its favorable performance especially in-case of handling large datasets which is not possible with the conventional SVM. Huang et al. \cite{huang2018twin} reviewed the research progress of TWSVM until 2017 for classification problems.

Support Vector Machine has also been used effectively for regression and is called support vector regression (SVR) \cite{basak2007support}. SVR is different than SVM in some aspects as it sets a margin of tolerance $\epsilon$ and find the optimal hyperplane such that the error is minimized. Thus, it finds a function such that error can be maximum of $\epsilon$ distance, thus any error within $\epsilon$ deviation is acceptable. Learning speed of SVR is low as it needs to solve a large sized QPP. Researchers have proposed many variants of SVR to improve its performance in terms of reducing computational complexity and improving accuracy \cite{drucker1997support, smola2004tutorial}. Some other variants of SVR includes $\epsilon$--support vector regression ($\epsilon$--SVR) \cite{shao2013varepsilon}, Fuzzy SVM \cite{chuang2007fuzzy}, $\nu$--SVR \cite{chang2002training}, robust and sparse SVR \cite{tanveer2016one} and some other algorithms \cite{yang2009localized,elattar2010electric}. $\epsilon$--SVR has high computational cost and in order to remediate this, an efficient twin support vector regression (TSVR) \cite{peng2010tsvr} determined two nonparallel hyperplanes that clusters the data points in two classes by optimizing the regression function. TSVR uses $\epsilon$--insensitive up and down-bound functions to optimize the final regression function. TSVR formulation is computationally less complex as it needs to solve a pair of smaller QPPs in lieu of solving one large QPP in SVR, thus it's speed is much more than SVR. However, in TSVR, the spirit of TWSVM was missing,   improved TSVR (TWSVR) \cite{khemchandani2016twsvr}  which works on the similar lines of TWSVM was proposed. To further boost the performance of TSVR, many algorithms have been introduced by researchers which are discussed in section 6.

In the past years, twin support vector classification and regression algorithms have been developed rapidly and are implemented to solve some real-life challenges. However, there is limited literature on twin support vector regression as it is a relatively new theory and needs further study and improvement. Thus, the objective of this paper is to present an compendium of recent developments in TWSVM and TSVR, identify limitations and advantages of these techniques and promote future developments. 

The framework of this paper is as follows, section \ref{sec:Related Work} presents brief about SVM and TWSVM, section \ref{sec:Research progress on twin support vector machines} and \ref{sec:Applications of Twin Support Vector Classification} include the recent advancements and applications of TWSVM respectively, section \ref{sec:Basic theory of Twin Support Vector Regression} presents a brief about TSVR, section \ref{sec:Research progress on Twin Support Vector Regression} and \ref{sec:Applications of Twin Support Vector Regression} include the recent advancements and applications of TSVR respectively and at last section \ref{sec:Future research and development prospects} provides synopsis, future research and development prospects.

\section{Related Work}
\label{sec:Related Work}
Suppose a binary classification problem with dataset $X \in \mathbb{R}^{l\times n}$  in which $l_1$ data points are in class $+1$ (termed as positive class) and $l_2$ data points are in class $-1$ (termed as negative class) and these are represented by matrix $A$ and $B$ respectively. For classification problems, the data label $Y\in\{-1,1\}$ and for regression $Y\in \mathbb{R}$. Accordingly, these matrices will be of size $(l_1 \times n)$ and $(l_2 \times n)$  where $n$ is feature space dimension and $l=l_1+l_2.$

\subsection{Support Vector Machine \cite{cortes1995support}}
The optimization problem of support vector machine is given as follows:
\begin{align}
\label{eqn:SVM}
    \underset{u,\xi}{min}~~&\frac{1}{2}u^Tu+c\sum_{k=1}^l \xi_k \nonumber \\
    s.t.~~ &y_k(u^T\phi(x_k)+b)\ge 1-\xi_k, ~~ k=1,\dots,l \nonumber \\
    &\xi_k\ge 0, k=1,\dots,l,
\end{align}
where, $c>0$ is the regularisation parameter, and $\phi(x)$ represents the non-linear mapping of the input $x$ and $\xi$ is the vector of slack variables.

Form the above given optimization problem, one can see that all the data samples appear in the constraints of the optimization problem. Hence, the complexity of the model is high as it involves solving a single large quadratic programming problem. Thus, the complexity of SVM is $O(l^3)$.

\subsection{Twin Support Vector Machines \cite{khemchandani2007twin}}
 Standard twin SVM is a binary classification model. To categorize data samples which cannot be separated by linear functions, TWSVM uses kernel functions to convert the higher dimensional data space to the required form. The two kernel generated surfaces are given as below:
\begin{align}
K(x^T,D^T)u_++b_+=0\,\,\mbox{and}\,\,K(x^T,D^T)u_-+b_-=0,
\end{align}
here $D=[A;B]$;  and $K$ is a kernel function. The formulation for nonlinear TWSVM classification is defined as below:
\begin{align}
\label{eqn:TWSVM1}
\min_{u_+,\,b_+,\,\xi_1}\hspace{.2em} &\frac{1}{2}\|K(A,\,D^T)u_++e_1b_+\|^2+{c_1}e_2^T\xi_1 \notag\\
s.t.\hspace{.4cm}  &-(K(B,\,D^T)u_++e_2b_+)+\xi_1\ge e_2,\,\,\xi_1\ge0
\end{align}
and 
\begin{align}
\label{eqn:TWSVM2}
\min_{u_-,\,b_-,\,\xi_2}\hspace{.2em} &\frac{1}{2}\|K(B,\,D^T)u_-+e_2b_-\|^2+{c_2}e_1^T\xi_2 \notag\\
s.t.\hspace{.4cm}&K(A,\,D^T)u_-+e_1b_-+\xi_2\ge e_1,\,\,\xi_2\ge0,
\end{align}
here $c_1, c_2 \geq 0, e_1, e_2$ are vectors of ones of suitable dimensions and $\xi_1, \xi_2$ are called slack variables. After using the Lagrange multipliers $\alpha\ge0, \beta\ge0$ and the Karush-Kuhn-Tucker (K.K.T.) \cite{kuhn} conditions, the duals of the QPPs in (\ref{eqn:TWSVM1}) and (\ref{eqn:TWSVM2}) are defined as below:
\begin{align}
\label{eqn:DTWSVM1}
\max_{\alpha}\hspace{.5em}& e_2^T\alpha-\frac{1}{2}\alpha^TQ(P^TP)^{-1}Q^T\alpha \notag\\
s.t.\hspace{.5em}&0\leq \alpha \leq c_1
\end{align}
and
\begin{align}
\label{eqn:DTWSVM2}
\max_{\beta}\hspace{.5em}& e_1^T\beta-\frac{1}{2}\beta^TP(Q^TQ)^{-1}P^T\beta \notag\\
s.t.\hspace{.5em}&0\leq \beta \leq c_2,
\end{align}
where $
   P =[K(A,\,D^T)\,\,e_1]\,\,\mbox{and}\,\, Q=[K(B,\,D^T)\,\,e_2].$
After  solving (\ref{eqn:DTWSVM1}) and (\ref{eqn:DTWSVM2}), the proximal hyperplanes are given as  follows:
\begin{align}
\begin{bmatrix}
u_+\\ b_+
\end{bmatrix} =-(P^TP+\delta I)^{-1}Q^T\alpha, 
\end{align}
\begin{align}
\begin{bmatrix}
u_-\\ b_-
\end{bmatrix} =(Q^TQ+\delta I)^{-1}P^T\beta, 
\end{align}
where $\delta I$ is a regularization term and $\delta>0.$
\par A new sample $x\in \mathbb{R}^n $ is allocated to either class on the basis of the proximity of kernel generated surface to $x$, i.e.,
\begin{align}
class(x)=\mbox{sign}\Big(\frac{{K(x^T,D^T)}u_++b_+}{\|u_+\|}+\frac{{K(x^T,D^T)}u_-+b_-}{\|u_-\|}\Big),
\end{align}
where $sign(\cdot)$ is signum function.

From optimization problems (\ref{eqn:TWSVM1}) and (\ref{eqn:TWSVM2}), one can see that the constraints of only one class appear in the optimization problem
 while generating the hyperplane for the other class. Thus, the size of the constraints in QPPs of TWSVM is approximately half compared to the SVM formulation (assuming the data is balanced between the classes). TWSVM is approximately four times faster than the
usual SVM. The complexity of TWSVM is $2*O((l/2)^3)$. In terms of generalization, TWSVM and SVM have comparable generalization performance \cite{khemchandani2007twin}.

\section{Research Progress on Twin Support Vector Machines}
\label{sec:Research progress on twin support vector machines}
In this section, we discuss the progress of TWSVM based models in classification problems. The variants of TWSVM are 
\subsection{Least Squares Twin Support Vector Machines}
\noindent
\newline
To reduce TWSVM training time, Kumar and Gopal \cite{kumar2009least} formulated least squares TWSVM (LS-TWSVM) algorithm. The major advantage of LS-TWSVM over TWSVM is that it only deals with linear equations in place of QPPs in TWSVM which reduces the computational complexity of the model. LS-TWSVM has comparable accuracy but low computational time than TWSVM. Although the computational time of LS-TWSVM is less than TWSVM, its generalization performance is poor as same penalties are allotted to the samples either being positive or negative. LS-TWSVM apply the empirical risk minimization (ERM) principle which affects accuracy and also causes over-fitting problem. Also, it doesn't take into account the effects of samples having different locations. To avoid this limitation,  weighted LS-TWSVM  \cite{xu2014weighted} gives different weights to the samples depending upon their locations. Experimental results have shown that weighted LS-TWSVM yields better testing accuracy than TWSVM and LS-TWSVM but its computationally time is more than these algorithms.

To incorporate expert's knowledge in LS-TWSVM classifier, knowledge-based LS-TWSVM (KBLS-TWSVM) \cite{kumar2010knowledge}  included polyhedral knowledge sets in the formulation of LS-TWSVM. Experimental results have shown that KBLS-TWSVM is simple and more apt classifier compared to LS-TWSVM. In order to improve accuracy of LS-TWSVM, Wang et al. \cite{wang2010localized} incorporated the manifold geometric structure of data of each class. It requires to solve a set of linear equations. 

Although, LS-TWSVM performed well with large datasets compared to TWSVM, Gao et al. \cite{gao20111} designed $l_1$-norm LS-TWSVM (NLS-TWSVM) to automatically select the relevant features in order to strengthen the algorithm in dealing with high-dimensional datasets. NLS-TWSVM is based on LS-TWSVM and includes a Tikhonov regularization term. To outdo LS-TWSVM in accuracy and avoid over-fitting problem, Improved LS-TWSVM (ILS-TWSVM)  \cite{xu2012improved} improved the classification accuracy by implementing structural risk minimization principle. ILS-TWSVM is faster and yields comparable generalization and computational time to LS-TWSVM. Since $L_2$ norm magnifies the outlier effect in least squares TWSVM models, hence, capped $L_{2,p}$ norm based least squares TWSVM \cite{yuan2021capped} was formulated to reduce the effect of outliers and noise.

To take advantage of the correlation between some data points and reduce the effect of noise, Ye et al. \cite{ye2012density} used density of the sample to allot weights for each sample (DWLSC). Experimental results demonstrate better classification accuracy of the DWLSC than TWSVM and LS-TWSVM.

There are several real-life problems which are essentially multicategory problems. The extensions of TWSVM to multicategory is also an active research domain where several research papers have evolved addressing the same. However, class imbalance problem is common in multicategory classification as all binary SVMs are trained with all patterns. To deal with these issues Nasiri et al. \cite{nasiri2014energy} formulated energy-based LS-TWSVM (ELS-TWSVM) in which the constraints of LS-TWSVM are converted to an energy model to decrease the impact of noise and outliers. TWSVM, LS-TWSVM, and ELS-TWSVM satisfy the empirical risk minimization (ERM) principle and also in these techniques the matrices are positive semi-definite. Thus, to remediate this problem, Tanveer et al. \cite{tanveer2016robust} embodied the SRM principle in ELS-TWSVM and proposed robust ELS-TWSVM (RELS-TSVM) which makes the matrices positive definite. Moreover, RELS-TSVM uses an energy parameter to handle the noise and thus make the algorithm more robust. 
Results have shown the promising generalization performance of RELS-TSVM with less computational time when compared with the baseline algorithms. In the recent comprehensive evaluation \cite{tanveer2019comprehensive} of $187$ classifiers on $90$ datasets, RELS-TSVM \cite{tanveer2016robust} emerged as the best classifier. Khemchandani and Sharma \cite{khemchandani2016robust} proposed robust least squares TWSVM (RLS-TWSVM)  which is fast and yields better generalization performance and is also robust to handle the noise. Further, the application in the field of human activity recognition is explored in the paper. The authors in \cite{khemchandani2016robust} also proposed Incremental RLS-TWSVM to increase the training speed of RLS-TWSVM and Incremental RLS-TWSVM also deals with noise and outliers effectively.

Tomar and Agarwal \cite{tomar2014feature} proposed feature selection based LS-TWSVM to diagnose heart diseases using F-score to choose the most relevant features which enhances the accuracy of the model. To handle imbalanced datasets,   a novel weighted LS-TWSVM \cite{tomar2014weighted} for imbalanced data which has better accuracy and geometric mean than SVM and TWSVM. To further enhance the accuracy on imbalanced datasets,  Hybrid Feature Selection Based WLS-TWSVM (HFS based WLS-TWSVM)  \cite{tomar2015hybrid} approach was proposed for diagnosing diseases like Breast Cancer, Diabetes and Hepatitis which can tackle data imbalance issues. Xu et al. \cite{xu2015structural} included data distribution information into the classifier structural LS-TWSVM (SLS-TWSVM). It is based on structural TWSVM and performs clustering before classification and embody the structural information into the model. SLS-TWSVM has a better generalization performance than $\nu$-SVM \cite{chen2005tutorial}, $\nu$-TWSVM \cite{peng2010nu}, STWSVM \cite{qi2013structural} and LS-TWSVM \cite{kumar2009least}. It also improves the noise insensitivity of LS-TWSVM.

Mei and Xu \cite{mei2019multi} proposed a novel multi-task LS-TWSVM algorithm which is build on directed multi-task TWSVM (DMTWSVM) and LS-TWSVM. It focuses on multitask learning instead of commonly applied single task learning in TWSVM and LS-TWSVM. This algorithm is computationally effective as it only requires to solve linear equations in lieu of QPPs in DMTWSVM.

Least square twin support vector hypersphere (LSTSVH)  \cite{tomar2015hybrid} is an enhancement of twin support vector hypersphere (TSVH) \cite{peng2013twin}. TSVH is different from TWSVM as it obtains two hyperspheres through solving two small SVM type problems. Experimental results demonstrate that LSTSVH has almost same accuracy as SVM, LS-SVM, TWSVM, LS-TWSVM, and TSVH but has high computational time than LS-SVM and LS-TWSVM and less than SVM and TWSVM. In 2020, Tanveer et al. \cite{tanveer2020largescale} proposed efficient large scale least squares twin support vector machines (LS-LSTWSVM) which uses different Lagrangian functions to eliminate the need for calculating the inverse matrices and thus enhance the performance of TWSVM on large scale datasets.

\begin{table}[h]
\caption {Classification accuracy on non-linear kernel \cite{tanveer2016robust}} \label{tab:title}
\resizebox{\textwidth}{!}{%
%\begin{tabular}{p{3.4cm} p{1cm}p{1.5cm}p{1.5cm}p{2cm}p{2cm}}
\begin{tabular} {l c c c c c }

\hline
Dataset (Train size, Test size) & TWSVM \cite{khemchandani2007twin}  & TBSVM \cite{shao2011improvements}  & LS-TWSVM \cite{kumar2009least}  & ELS-TWSVM \cite{nasiri2014energy} & RELS-TSVM \cite{tanveer2016robust}\\
\hline
Ripley (250$\times$2, 1000$\times$2) & 88.00 & 83.50 & 86.40 & 83.50 & 88.30 \\
\hline
Heart-c (177$\times$13, 120$\times$13) & 60.81 &  70.00 & 61.67 & 61.67 & 67.50 \\
\hline
Heart-stat (200$\times$13, 70$\times$13) &  82.85 & 84.28 & 85.71 & 78.57 & 85.71\\
\hline
Ionosphere (246$\times$33, 105$\times$33) & 92.38 & 87.46 & 87.61 & 92.38 & 96.19\\
\hline
Bupa-liver (241$\times$6, 104$\times$6)& 63.46 & 70.19 & 55.77 & 66.35 & 69.23 \\
\hline
Votes (306$\times$16, 129$\times$16)& 96.90 & 96.90 & 96.12 & 95.35 & 96.90\\
\hline
WPBC (137$\times$34, 57$\times$34)& 80.70 & 80.19 & 78.94 & 75.44 & 80.70\\
\hline
Pima-Indian (537$\times$8, 231$\times$8)& 77.48 & 76.09 & 64.07 & 78.35 & 80.52
 \\
\hline
\end{tabular}}
\end{table}

\begin{figure}
     \centering
     \includegraphics[width=\linewidth]{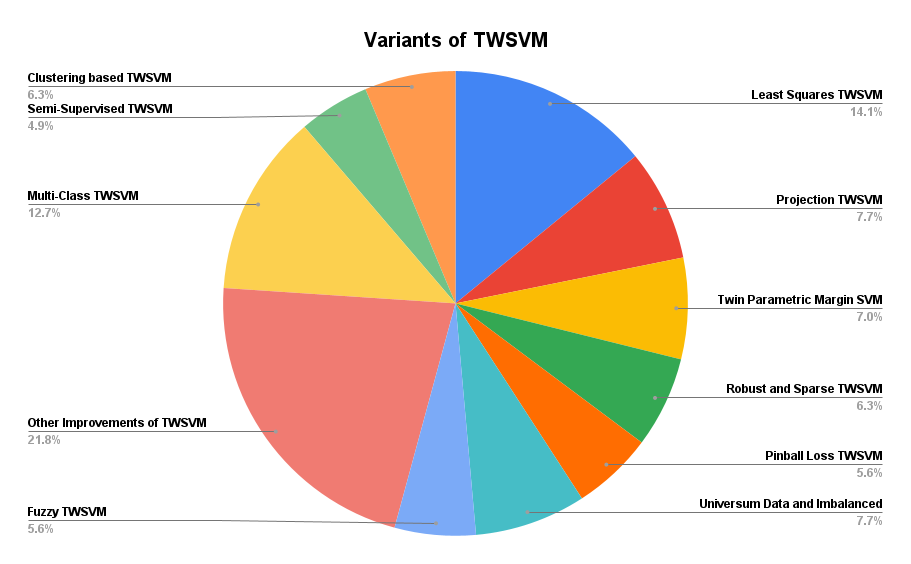}
     \caption{Variants of TWSVM}
     \label{fig:Variants of TWSVM}
 \end{figure}

\subsection{Projection Twin Support Vector Machine}
\noindent
\newline
Projection TWSVM (P-TWSVM) \cite{chen2011recursive} is a multiple-surface classification (MSC) technique based on the multi-weight vector projection SVM (MVSVM) and TWSVM. In 2011, Chen et al. \cite{chen2011recursive} introduced P-TWSVM with minimizing within class variance principle. Unlike finding hyperplanes in TWSVM, P-TWSVM finds projection axes to distinctly separate samples. To enhance performance of P-TWSVM, authors proposed a recursive algorithm in which for each class multiple projection axes are considered. P-TWSVM is a better classifier as hyperplanes are more fitted for single plane based samples while for datasets having complex sample distribution like XOR problems, projection directions can generate better accuracy. Experimental results have shown that P-TWSVM has better accuracy than GEPSVM, TWSVM, LS-TWSVM, and MVSVM. However, it is computationally more complex than TWSVM. Thus, Shao et al. \cite {shao2012least} proposed improved P-TWSVM and used the idea in LS-SVM \cite{suykens1999least} and LS-TWSVM \cite{kumar2010knowledge} and added a regularization term into P-TWSVM simultaneously maintaining the optimization problem to be positive definite. It is simple, fast and has similar classification accuracy but less computational time than P-TWSVM. 

Although P-TWSVM is an efficient classifier, it only implements empirical risk minimization principle which can incur possible singularity problems which need to be dealt with principal component analysis (PCA) and linear discriminant analysis (LDA). Shao et al. \cite{shao2013regularization} formulated a new P-TWSVM variant by introducing a maximum margin regularization term called RP-TWSVM in which empirical risk is replaced by regularized risk. Authors further proposed a successive over-relaxation (SOR) technique and a genetic algorithm (GA) for parameter selection to optimally solve the primal problems. RP-TWSVM not only has better accuracy but also has a better generalization than P-TWSVM.

To implement least square P-TWSVM to non-linear problems, Ding and Hua \cite{ding2014recursive} introduced kernel to LSP-TWSVM \cite{shao2012least} which also deals with linear equations similar to LSP-TWSVM and opposed to P-TWSVM. Authors introduced nonlinear recursive algorithm in order to improve its performance. Experimental results have shown good classification accuracy than LSP-TWSVM and P-TWSVM.

Guo et al. \cite{guo2014feature} proposed a feature selection approach for LSP-TWSVM which finds two projection directions and has comparable prediction accuracy to that of TWSVM, LS-TWSVM and LSP-TWSVM and a similar generalization ability to TWSVM and LS-TWSVM.

LSP-TWSVM enhances the performance of P-TWSVM but fails to include underlying data information to enhance classification accuracy. To overcome this limitation, weighted LSP-TWSVM (LIWLSP-TWSVM)  \cite{hua2015weighted}  exploited local information into the algorithm. LIWLSP-TWSVM is more effective than LSP-TWSVM as it uses weighted mean rather than standard mean in LSP-TWSVM. It has better generalization ability but its computational cost is high when solving multi classification problems.

Hua et al. \cite{hua2017novel} formulated a novel projection TWSVM (NP-TWSVM). NP-TWSVM has many advantages over P-TWSVM. It is faster than P-TWSVM as calculation of inverse matrices is not avoided in NP-TWSVM. It determines the projection axes so that margin across the sample and class is greater than or equal to zero. Experimental results have shown that it obtains better generalization and accuracy when compared with other baseline algorithms. An efficient nonparallel sparse projection SVM (NPrSVM)  \cite{chen2020nprsvm}   finds optimal projection for each class such that the projection values of within class instance are clustered as much as possible within an insensitive tube while those of other class instance are kept away. Extensive experiments show that NPrSVM is superior than PTWSVM in terms of generalization performance, training time, sparsity and robustness to outliers. $\nu-$ projection twin SVM \cite{chen2020nu} seeks projection axis for each class in a manner that $\nu$ controls the fraction of support vectors and error margin, and also avoids the matrix inversions. Robust rescaled hinge loss based projection TWSVM model \cite{ren2021robust} use different parameters to control the effect of outliers and the model results in better performance.

\begin{table}[h]
\caption {Classification accuracy on non-linear kernel \cite{hua2015weighted}} \label{tab:title}
\resizebox{\textwidth}{!}{%
%\begin{tabular}{p{3.3cm} p{1.8cm}p{2.4cm}p{2.5cm}p{2.5cm}p{3.2cm}}
\begin{tabular} {l c c c c c }

\hline
Dataset & TWSVM\cite{khemchandani2007twin}
& WLTSVM\cite{ye2012weighted} & P-TWSVM\cite{chen2011recursive} & LSP-TWSVM\cite{shao2012least} & LIWLSP-TWSVM\cite{hua2015weighted} \\
\hline
Cleve (296$\times$13) & 85.12 & 84.90 & 85.93 & 83.80& 85.30\\
\hline
Bupa-liver (345$\times$6) & 74.13 & 73.58& 72.18 & 72.47& 73.39  \\
\hline
Heart (270$\times$13) & 84.44 & 84.82 & 84.44& 84.44 & 84.82\\
\hline
Monks2 (432$\times$6) & 68.77 & 66.21 & 68.71& 67.84 & 68.95\\
\hline
Australian (690$\times$14) & 86.97 & 86.70 & 86.99 & 87.27 & 88.13\\
\hline
Wpbc (198$\times$33) & 79.36 & 80.35& 78.03 & 78.56 & 81.87\\
    
\hline
Parkinsons (195$\times$22) & 87.22 & 85.00 & 86.36 & 86.62 & 87.48\\
\hline

Vertebral (310$\times$6) & 85.81 & 86.13 & 84.52& 84.84& 87.10
\\
\hline

Spect (267$\times$23) & 84.42  & 85.19 & 84.42& 84.80 &85.11\\
\hline

Spectf (267$\times$44) & 83.43 & 82.96 & 82.49 & 83.26 & 84.25\\
\hline
\end{tabular}}
\end{table}

\subsection{Twin Parametric Margin Support Vector Machine}
\noindent
\newline
Motivated by the idea of TWSVM and Par-$\nu$-SVM \cite{hao2010new}, in 2011 Peng et al. \cite{peng2011tpmsvm} formulated twin parametric-margin SVM (TPMSVM). Like par-$\nu$-SVM, TPMSVM seeks to determine two parametric margin hyperplanes which defines the positive and negative margin respectively. It's an indirect classifier even suitable when noise is heteroscedastic as it automatically adjusts a flexible margin. Experimental results have shown that TPMSVM has comparable generalization ability to SVM, Par-$\nu$-SVM, and TWSVM. Also, TPMSVM has much lower training time than Par-$\nu$-SVM as it solves two small sized QPPs like TWSVM. TPMSVM classifier is not sparse due to the formulations of its parametric- margin hyperplanes. TPMSVM has good generalization but it is computationally complex as it solves two QPPs. In 2013, Wang et al. \cite{wang2013ga} added a quadratic function to TPMSVM in primal space to get better training speed. The proposed algorithm is called smooth twin parametric-margin SVM (STPMSVM). Also, the authors proposed the use of genetic algorithm (GA) in STPMSVM to  overcome the drawback of regularizing at least four parameters in TPMSVM.  To obtain sparsity and feature noise insensitivity, truncated pinball loss TPMSVM \cite{wang2021twin} was proposed. The optimal separating hyperplanes are obtained via concave–convex procedure (CCCP).

Shao et al. \cite{shao2013least} in 2013 proposed least squares TPMSVM (LSTPMSVM) to decrease the training cost of TPMSVM as LSTPMSVM solves two primal problems rather than dual problems. This change makes it less complex and increases training speed. It has comparable or better classification accuracy but with remarkably less computational time than TPMSVM. For model selection, the authors proposed a particle swarm optimizer (PSO) which effectively optimizes the four parameters defined in LSTPMSVM. In terms of generalization, LSTPMSVM performs better than TPMSVM and LS-TWSVM. In order to consider prior structural data information, Peng et al. \cite{peng2013structural} in 2013 proposed structural twin parametric-margin SVM (STPMSVM). Based on cluster granularity, the class data structures are included in the problem. STPMSVM not only obtained good generalization ability but also showed fast learning speed, and better performance than TPMSVM. TPMSVM is an efficient classifier but it losses the sparsity due to the weight vectors of the hyperplanes. To solve this issue, centroid-based TPSVM (CTPSVM)  \cite{peng2015improvements} was introduced which uses the projection of the centroid points and leads to a sparse optimal hyperplane by optimizing the centroid projection values.

To incorporate structural information present in data, in 2017, Rastogi et al. \cite{rastogi2018robust} proposed robust parametric TWSVM for pattern classification (RP-TWSVM) which seek to find two parametric margin hyperplanes that has the capability to adjust margin to capture heteroscedastic noise data information. Rastogi et al. \cite{rastogi2018anglea},  formulated angle-based TPSVM (ATPSVM) is proposed, which can efficiently handle heteroscedastic noise. Other improvements which maximizes angle between the twin hyperplanes are proposed by Rastogi et al. in \cite{rastogi2018angle}. Taking motivation from TPMSVM, ATPSVM determines a pair of hyperplanes so that the angle between their normals is maximized. Recently, Richhariya and Tanveer proposed a novel angle based universum LS-TWSVM (AULSTWSVM) for pattern classification. In contrast to ATPSVM \cite{rastogi2018angle}, the AULSTWSVM minimizes the angle between universum hyperplane and classifying hyperplane.  To incorporate the prior information about the distribution of the data using the universum, AULSTWSVM used linear loss \cite{shao2015weighted} in the formulation. Numerical experiments show the promising generalization performance with very less computational time. Further, the application in the diagnosis of Alzheimer's disease is explored.

Moreover, the proposed AULSTWSVM includes linear loss \cite{shao2015weighted} in the optimization problem, while incorporating prior information about data distribution using the universum. Moreover, the solution of AULSTWSVM involves system of linear equations leading to less computation time \cite{kumar2009least}

\subsection{Robust and Sparse Twin Support Vector Machine}\noindent
\newline
TWSVM achieves better accuracy and is faster  when compared to conventional SVM but it loses sparsity as it considers $l_2$ norm of distances in the objective function. Thus, in order to make TWSVM sparse, Peng \cite{peng2011building} in 2011 proposed TWSVM in primal space which provides a sparse hyperplane with better generalization ability. To improve the robustness, a regularization term is also added. It has comparable generalization and a rapid learning speed to TWSVM and LS-TWSVM but the computational cost is high. In 2013, Peng and Xu \cite{peng2013robust} proposed robust minimum class variance TWSVM (RMCV-TWSVM) classifier to enhance generalization and robustness of TWSVM. This algorithm has an extra regularization term which makes its learning speed comparable to TWSVM but obtains better generalization ability than TWSVM.

Qi et al. \cite{qi2013robust} in 2013 proposed robust TWSVM using second-order cone programming formulation. It is effective with noise-corrupted data. This algorithm overcomes the limitation of TWSVM and  computes the inverse of matrices which are not suitable for large datasets, it takes only the inner products about samples by which kernel trick can be applied directly and does not need to solve the extra inverse of matrices. It is superior in computational time and accuracy than other TWSVMs. In 2014, Tian et al. \cite{tian2014efficient} proposed sparse nonparallel support vector machine (SN-SVM). While TWSVM losses the sparseness, SN-SVM has the inherent sparseness as it uses the two loss functions instead of one in existing TWSVM. Only the empirical risk is considered in TWSVM, SN-SVM introduces the regularization term by maximizing the margin. TWSVM minimizes the loss function based on $l_1$ or $l_2$-norm. Thus, Zhang et al. \cite{zhang2014sparse} proposed $l_p$-norm-LS-TWSVM which is an adaptive as $p$ can be automatically chosen by data. Robust non-parallel SVM via second order cone programming \cite{lopez2019robust} is robust to outliers and noise, and constructs the two separating hyperplanes via maximisation of probabilistic framework. 

In order to intensify the robustness and sparsity in the original formulation of TWSVM, Tanveer \cite {tanveer2015robust} in 2015, incorporated regularization technique and formulated a linear programming $l_1$-norm TWSVM (NLP-TWSVM) which needs to solve linear equations rather than solving QPPs in TWSVM which makes it fast, robust, sparse and a simple algorithm. Experimental results demonstrate that NLP-TWSVM's generalization ability is better and computational time is less than GEPSVM, SVM, and TWSVM. Tanveer \cite{tanveer2015application,tanveer2013smoothing} proposed smoothing approaches for 1-norm linear programming TWSVM (SLPTWSVM). The solution of SLPTWSVM reduces to solving two systems of linear equations which makes the algorithm extremely simple and computationally efficient. Tanveer et al. \cite{tanveer2019sparse} proposed sparse pinball TWSVM (SP-TWSVM) by introducing \(\epsilon\) insensitive zone pinball loss function in the orginal TWSVM formulation. SP-TWSVM is noise insensitive, retain sparsity and more stable for re-sampling. Robust capped $l_1$ norm TWSVM \cite{wang2019robust} reduced the effect of outliers which resulted in better performance. To reduce the overfitting issues, capped $l_1$ norm twin bounded support vector machines \cite{ma2020capped} was proposed.  Efficient and robust $l_1$ norm TWSVM \cite{yan2019efficient} used $l_1$ norm to maximise the ratio of interclass scatter to intraclass scatter. The authors used iterative procedure to get the optimal separating hyperplanes. Adaptive capped $L_{\theta,\epsilon}$-loss based TWSVM \cite{ma2021adaptive} is a generalized TWSVM model wherein $\theta$ and $\epsilon$ parameters are optimized to meet the objectives of robust to noise and outliers.

\subsection{Pinball Loss Twin Support Vector Machine}
\noindent
\newline
In 2016, Xu et al. \cite{xu2016novel} proposed a novel TPMSVM with the pinball loss (Pin-TPMSVM). The authors introduced pinball loss function that is based on maximizing quantile distances between the two classes instead of hinge loss function which maximizes the distance between the closest samples of the two classes. This leads to noise insensitivity and as well as re-sampling stability. Pin-TWSVM has excellent capability to handle noise which makes the model a more robust classifier but little attention is given on sparsity due to which the testing time is high and there is instability in re-sampling.  Xu et al.  \cite{xu2016maximum} also formulated a maximum margin and minimum volume hyperspheres with pinball loss (Pin-$M^3$HM). Authors proposed this algorithm to enhance the generalization of twin hypersphere SVM (TH-SVM) for noise present in datasets. The algorithm classifies samples of two classes by a pair of hyper-spheres, each containing either majority or the minority class samples simultaneously maintaining maximum margin between the hyperplanes. Pin-$M^3$HM is fast and has better accuracy than TWSVM. In 2018, Sharma and Rastogi \cite{sharma2018insensitive} proposed two models called $\epsilon$-Pin-TWSVM and Flex-Pin-TWSVM. $\epsilon$-Pin-TWSVM introduces $\epsilon$ parameter to reduce the impact of noise and to attain a sparse solution while Flex-Pin-TWSVM uses a self-optimized framework which makes this algorithm flexible to estimate the size of insensitive zone. It adapts to the structure of data. Pin-TPMSVM has impressive ability to handle noise but to reduce parameter tuning time, Yang et al. \cite{yang2018piecewise} in 2018 introduced a new solution approach for the TPMSVM in which one instance is considered at a time and it is solved analytically without solving optimization problem. It is fast, simple and flexible.  Since pinball loss function is not differentiable at zero, hence, smooth pinball loss TWSVM \cite{li2021smooth} used smooth approximation function and solved the objective functions via Newton-Armijo method.

For imbalanced data classification in 2018, Xu et al. \cite{xu2018maximum} proposed a maximum margin twin spheres which uses pinball loss. This algorithm finds homocentric spheres so that the smaller one captures positive class samples and the larger one repel negative samples. It requires to solve a QPP and a Linear programming (LP) problem. This algorithm has good generalization performance and noise insensitivity,  however, suffers due to large complexity. To reduce the complexity, bound estimation-based safe acceleration for maximum margin of twin spheres machine with pinball loss \cite{yuan2021bound} was proposed. 
In 2019, Sharma et al. \cite{sharma2019large} proposed a Stochastic Quasi-Netwon method based TPSVM using Pinball Loss Function (SQN-PTWSVM) which can scale the training process to handle millions of data points while simultaneously deals with noise and re-sampling data issues. Twin neural networks \cite{pant2019twin} uses  feature maps which allows better discrimination among classes. The twin neural network is also extended to multiclass problems wherein the number of neural networks trained is proportional to number of classes.

The aforementioned pinball loss TWSVM algorithms are based on TPMSVM and not on the original TWSVM algorithms. In 2019, Tanveer et al. \cite{tanveer2019general} introduced pinball loss to the original TWSVM, termed as general TWSVM with pinball loss (Pin-GTSVM). Pin-GTSVM \cite{tanveer2019general,ganaie2021robust} is less sensitive to the outliers and more stable algorithm for re-sampling as compared to the original TWSVM. To retain the sparsity of original TWSVM and Pin-GTSVM, Tanveer et al. \cite{tanveer2019sparse} proposed a novel sparse pinball TWSVM (SP-TWSVM) which uses \(\epsilon\)-insensitive zone pinball loss function. SP-TWSVM has better classification accuracy than TWSVM and Sparse Pin SVM and it is insensitive to outliers, retains sparseness and suitable for re-sampling. Recently, Tanveer et al. \cite{tanveer2019improved} proposed improved sparse pinball TWSVM (ISPTWSVM) by adding a regularization term to the objective function of SP-TWSVM. ISPTWSVM implements SRM principle and also the matrices appear in the dual formulation are positive definite which makes the proposed algorithm computationally less complex and it also achieves better accuracy than other baseline algorithms. Most of the twin SVM based models involve matrix inversion operations which limits their applicability to large scale data. Hence, large scale pinball TWSVM \cite{tanveer2021large} uses pinball loss function to reduce the issues of feature noise and reformulated the Lagrangian in a manner that matrix inversions are no longer involved in the optimization problems. This scaled the model to large scale data.

\subsection{Twin Support Vector Machine for Universum Data and Imbalanced Datasets}
\noindent
\newline
In 2012, Qi et al. \cite{qi2012twin} formulated TWSVM for universum data classification (U-TWSVM). Universum data is defined as the data samples which does not belong to any given class. This algorithm utilizes the universum data to enhance TWSVM classification accuracy. U-TWSVM employs new data points to the either  class based on its proximity to the hyperplanes. Experimental results have shown that U-TWSVM has better accuracy than TWSVM. In order to construct a robust classifier by including the prior information embedded in the universum samples in 2014, Qi et al. \cite{qi2014nonparallel} formulated nonparallel SVM (U-NSVM) which maximizes the two margins related to the two nearest adjacent classes.

As many real life problems consists of datasets which are imbalanced in nature i.e. classes do not contain same number of data samples, due to which many machine learning algorithms cannot be implemented. Thus, to enhance the performance of TWSVM while dealing with imbalanced datasets, in 2014 Shao et al. \cite{shao2014efficient} proposed weighted Lagrangian TWSVM for imbalanced data classification. Authors introduced a quadratic loss function to the formulation of TWSVM which enabled faster training of data points. Also, this is more robust to outliers and has better computational speed and accuracy than TWSVM. In 2014, Tomar et al. \cite{tomar2014weighted} proposed weighted LS-TWSVM for imbalanced datasets by adjusting the classifier and assigning distinct weight to training samples. The results of the experiment shows that its accuracy is more than SVM, TWSVM, and LS-TWSVM. Furthermore, in 2015, Xu et al. \cite{xu2016least} formulated least squares TWSVM (ULS-TWSVM) which exploits the universum data. It introduces a regularization term thus implements SRM principle and is computationally less complex as it requires to solve linear equations in place of QPPs in TWSVM. Cao and Shen \cite{cao2016combining} proposed combining re-sampling with TWSVM for imbalanced data classification. In this algorithm, authors presented a hybrid re-sampling technique which utilizes the one side selection (OSS) algorithm and synthetic minority oversampling technique (SMOTE) algorithm to balance the training data. This is combined with TWSVM for classification purpose. However, SMOTE algorithm based on $K$-nearest neighbors can often result in over-fitting. Robust rescaled hinge loss TWSVM \cite{huang2019robust} and density weighted TWSVM \cite{hazarika2021density}  for imbalanced datasets resulted in improved performance.

Encouraged by the performance of $\nu$-TWSVM \cite{peng2010nu} and $U$-TWSVM \cite{qi2012twin}, Xu et al. \cite{xu2016nu} in 2016 proposed $\nu$-TWSVM for universum data classification ($U\nu$-TWSVM). It is more flexible in using the prior knowledge from universum data to improve the generalization ability. It uses two hinge loss functions so that data can remain in a nonparallel insensitive loss tube. Experimental results have shown that $U\nu$-TWSVM has better accuracy and also costs lower running time than other baseline algorithms. 

In 2017, Xu \cite{xu2017maximum} proposed maximum margin twin spheres SVM algorithm (MMTSSVM) to overcome many limitations of TWSVM while dealing with imbalanced data. This algorithm seeks to determine two homocentric spheres so that smaller one captures as many positive class samples and the larger one repel negative samples simultaneously increases the margin between the spheres. 

In 2018, Richhariya et al. \cite{richhariya2018improved} added a regularization term into the optimization problem of universum TWSVM to make matrices non-singular and improve the generalization performance. The proposed algorithm is called improved universum TWSVM. It has better generalization and training time than USVM and UTWSVM. In 2019, Richhariya and Tanveer \cite{richhariya2019fuzzy} proposed a fuzzy universum SVM (FUSVM) based on information entropy. Universum support vector machines are more efficient than other SVM methods as it includes prior information of samples. But, it is not effective in case of noise-corrupted datasets. FUSVM introduces weights such that it gives fewer weights to the outlier universum points. Further, the authors proposed a fuzzy universum TWSVM (FUTWSVM). Both the algorithms have better generalization performance than SVM, USVM, TWSVM, and UTWSVM. Recently, Richhariya and Tanveer \cite {richhariya2020reduced} proposed a reduced universum twin support vector machine for class imbalance learning (RUTWSVM-CIL) with the idea of prior information about the data distribution. RUTWSVM-CIL used reduced kernel matrix, and thus applicable for the large sized imbalanced datasets. Numerical experiments on several imbalanced benchmark datasets showed the applicability of RUTWSVM-CIL. In 2020, Richhariya and Tanveer \cite{richhariya2020universum} also proposed a novel parametric model for universum based twin  support  vector machine and extended its application for the classification of Alzheimer's disease data. Fuzzy universum least squares TWSVM \cite{richhariya2021fuzzy,borah2018improved} and fuzzy least squares  TWSVM \cite{ganaie2021fuzzy,gupta2018entropy} solved a linear system of equations instead of solving the QPPs, hence, the model is fast and requires no external toolbox for solving the optimization problems. Robust  TBSVM \cite{borah2021robust} is proposed to make the model more robust to noise in imbalance datasets.

\subsection{Fuzzy Twin Support Vector Machines}
In 2017, Chen and Wu \cite{chen2018new} proposed a novel fuzzy TWSVM (NFTWSVM) which takes care of the problem of dealing with one class playing major role in classification, by assigning fuzzy membership to different samples. The proposed NFTWSVM includes fuzzy neural networks and provides more generalized results. In 2019, Richhariya and Tanveer \cite{richhariya2019fuzzy} proposed a fuzzy universum SVM (FUSVM) based on information entropy. This algorithm is also discussed in Section 3.6. Richhariya and Tanveer \cite{richhariya2018robust} also proposed a robust fuzzy LS-TWSVM (RFLS-TWSVM-CIL) to boost the performance of TWSVM on imbalanced datasets. The optimization problem in this algorithm is strictly convex as it uses 2-norm of the slack variables. Authors further proposed a fuzzy membership function to deal with imbalanced problems as it provides weights to samples and includes data imbalanced ratio. RFLS-TWSVM-CIL obtains better generalization and is computationally more efficient as it requires to solve two linear equations. However, RFLSTWSVM-CIL algorithm implements empirical risk minimization principle which requires inverse of matrices to be positive semi-definite. In order to improvise, Ganaie et al. \cite{ganaie2020regularized} introduced a regularization term to the primal formulation of RFLSTWSVM-CIL. The proposed algorithm is regularized robust fuzzy least squares (RRFLSTWSVM) which doesn't require the extra assumption of inverse of matrices to be positive semi-definite and performs better than RFLSTWSVM-CIL.

In 2018, Khemchandani et al. \cite{khemchandani2018fuzzy} formulated fuzzy LS version of TSVC in which each data point has a fuzzy membership value and is allotted to different clusters. This algorithm is also discussed in Section 3.11. In 2020, Chen et al. \cite{chen2020entropy} proposed entropy-based fuzzy least squares TWSVM which considers the fuzzy membership for each data point basis entropy. Rezvani and Wang \cite{rezvani2019intuitionistic} formulated intuitionistic fuzzy TWSVM (IFTWSVM) which is modified version of fuzzy TWSVM as it considers the position of input data in the feature space and calculate adequate fuzzy membership and also reduces the effect of noise. Zhang and Li \cite{zhang2019fuzzy} proposed fuzzy TWSVM which assigns fuzzy membership based on intra-class hyperplane and sample affinity. Experimental results showed that this algorithm has better accuracy than TWSVM and classic fuzzy TWSVM. In 2019, Gupta et al. \cite{gupta2019fuzzy} proposed entropy-based fuzzy twin support vector
machine (EFTWSVM-CIL) which is an efficient algorithm for imbalanced datasets as it assigns fuzzy membership values based on entropy of the samples. In 2020, Chen at el. \cite{chen2020entropy} proposed entropy-based fuzzy least squares twin support vector machine (EFLSTWSVM) which is an improvised version of EFTWSVM-CIL by formulating the QPPs in least squares sense and retains the superior characteristics of LSTWSVM. Experimental results shown that this algorithm performed better than other baseline fuzzy TWSVM algorithms.

\begin{table}[h]
\caption {Classification accuracy on non-linear kernel \cite{chen2020entropy}} \label{tab:title}
\resizebox{\textwidth}{!}{%
%\begin{tabular}{p{4.3cm} p{2cm}p{2.2cm}p{2cm}p{2cm}p{3.0cm}p{2.8cm} }
\begin{tabular}{l c c c c c c }

\hline
Dataset  (Train size, Test size)& TWSVM \cite{khemchandani2007twin} & LSTWSVM \cite{kumar2009least} & EFSVM \cite{fan2017entropy} & FTWSVM \cite{li2013fuzzy}& EFTWSVM-CIL \cite{gupta2019fuzzy} & EFLSTWSVM \cite{chen2020entropy}\\

\hline
Australian (690$\times$14) & 87.83  & 87.54  & 87.10  & 87.39  & 87.68  & 87.97 \\

\hline
Bupa-Liver (345$\times$6) & 75.07  & 75.94  & 73.91  & 74.78  & 74.49  & 75.65 \\

\hline
House-Votes (435$\times$16) & 96.32  & 96.55  & 95.86  & 96.32  & 96.09  & 
96.55 \\

\hline
Heart-c (303$\times$13) & 85.14  & 84.50  & 84.84  & 83.83  & 86.13  & 85.82  \\

\hline
Heart-Statlog (270$\times$13) & 85.56  & 85.19  & 85.19  & 85.86  & 86.30  & 
85.93  \\

\hline
Ionosphere (351$\times$34) & 94.32  & 94.30  & 95.73  & 94.60  & 94.60  & 96.58  \\

\hline
Musk (476$\times$166) & 95.79  & 94.96  & 94.54  & 96.43  & 95.58  & 96.63 \\

\hline
PimaIndian (768$\times$8) & 78.52  & 78.30  & 77.74  & 78.13  & 78.64  & 79.16  \\

\hline
Sonar (208$\times$60) & 88.95  & 91.39  &
90.88  & 89.94  & 91.86  & 91.81  \\

\hline
Spect (267$\times$44) & 82.42  & 83.89 & 82.78  & 83.15  & 83.52  & 84.65  \\

\hline
Wpbc (198$\times$34) & 83.86  & 83.36  & 82.83  &  83.33  & 84.36  & 84.88  \\

\hline
\end{tabular}}
\end{table}

\subsection{Some other improvements of Twin Support Vector Machines}

Kumar and Gopal \cite{kumar2008application} in 2008 enhanced TWSVM using smoothing techniques. Authors proposed to transform the primal problems of TWSVM into smooth unconstrained minimization problems and used Newton$-$Armijo algorithm for optimization. Smooth TWSVM (STWSVM) has comparable generalization to TWSVM but is significantly a faster algorithm. Jayadeva et al. \cite{khemchandani2009optimal} in 2009 proposed iterative alternating algorithm to make kernels learn efficiently. In 2009, Zhang \cite{zhang2009boosting} proposed boosting based TWSVM for clustered microcalcifications detection. Results have shown that this method improved detection accuracy. Bagging algorithm was combined with boosting to solve the unstable problem of TWSVM. This method is called BB-TWSVM. Twin support vector machine in linear programs and robust TWSVM \cite{TWSVM13} have also been proposed for the classification problems \cite{li2014twin}.

In 2010, Shao et al. \cite{shao2010multiple} proposed a bi-level programming method to multiple instance classification, called MI-TWSVM. Multiple instance learning (MIL) is a supervised technique wherein the training data points consist of labeled bags and these bags include multiple unlabeled instances. For binary classification, a bag is a negative sample if all the instances in it are negative and it is labeled a positive sample if at least one instance is positive.
Similar to TWSVM, the proposed algorithm seeks two hyperplanes thereby the positive hyperplane is closest to the positive instances and  as distant as possible from the negative instances and vice-versa. In 2010, Ghorai et al. \cite{ghorai2010unity} proposed a unity norm TWSVM classifier (UNTWSVM) which includes unity norm equality constraints and a quadratic loss function in the objective function of TWSVM. This algorithm can be solved by sequential quadratic optimization method. UNTWSVM has more computational cost than TWSVM especially for large datasets because of the nonlinear formulation.  Generalized TWSVM \cite{moosaei2021generalized} proposed two models, in the first model the authors used the $l_1$ and $l_\infty$ norm in the optimization problems and  in second model the convex, piecewise quadratic objective function is solved via generalized Newton method.

In order to improve TWSVM's generalization ability and to decrease support vectors (SVs), in 2010 Peng \cite{peng2010nu} introduced a variable $p$ and parameter $\nu$ in TWSVM and proposed $\nu$-TWSVM. An improved $\nu$-TWSVM \cite{xu2014improved} has also been proposed for the classification problems.  The parameter $\nu$ controls the trade-off between the SV and marginal error while adaptive $p$ overcomes the limitations of constraints being over-restricted in TWSVM and thus reduces the number of support vectors. But, it gives the same penalties to each misclassified sample and results in over-fitting. Thus, Xu et al.  \cite{xu2012rough} in 2012 introduced the rough set theory into $\nu$-TWSVM to remediate this limitation. This algorithm is more effective in avoiding over-fitting as it gives different penalties to different points depending upon location. It also has better generalization than $\nu$-TWSVM. 

Although rough $\nu$-TWSVM \cite{xu2012rough} performs better than $\nu$-TWSVM, it provides different weights to negative samples and same weights to all the positive samples. Thus, Xu et al. \cite{xu2014knn} formulated K nearest neighbor (KNN) weighted rough $\nu$-TWSVM which gives different penalties to the samples of both the classes. It has better accuracy and lower computational cost than $\nu$-TWSVM and Rough $\nu$-TWSVM. In another attempt to enhance the results of $\nu$-TWSVM, Khemchandani et al. \cite{khemchandani2016improvements} formulated two models for binary classification: $I\nu$TWSVM and $I\nu$TWSVM Fast. Both the algorithms are faster than $\nu$-TWSVM as $I\nu$TWSVM solves one small QPP and an unconstrained minimization problem (UMP) while $I\nu$TWSVM Fast solves one unimodal function and one UMP.  To overcome the disadvantage of TWSVM being sensitive to outliers, Xie and Sun \cite{xie2015multitask} implemented class centroid and proposed multitask centroid TWSVM for multitask problems.

For handling large scale datasets, Shao et al. \cite{shao2015weighted} introduce weighted linear loss in TWSVM and proposed weighted linear loss TWSVM (WL-TWSVM) to classify large scale datasets. The proposed algorithm only deals with linear equations which increases the training speed and also has better generalization than TWSVM.  To further fasten up the training process, Sharma et al. \cite{sharma2018stochastic} and Wang et al. \cite{wang2018insensitive}, recently proposed stochastic conjugate gradient method based twin support vector machine (SCH-TWSVM). The resulting model showed effective performance on binary activity classification problem. 

Shao et al. \cite{shao2011improvements} in 2011  introduced the regularization term in TWSVM to embody SRM principle and formulated a new algorithm called twin bounded support vector machines (TBSVM). To further speed up training, SOR technique is used. Numerical results on various datasets show that TBSVM is faster and has better generalization ability than TWSVM. In 2010, Ye et al. \cite{ye2011localized} formulated localized TWSVM via convex minimization (LC-TWSVM) which effectively constructed two nearest-neighbor graphs in the original input space to reduce the space complexity of TWSVM. Shao and Deng \cite{shao2013novel} in 2012 considered the unity norm constraints and added a regularization term to minimize structural risk in TWSVM and proposed margin-based TWSVM with unity norm hyperplanes (UNHMTWSVM). The proposed algorithm is fast, has better generalization and accurate  compared to TWSVM. Lagrangian TBSVM with $L_2$ norm \cite{gupta2019lagrangian} replaced $L_1$ norm of the slack variables with $L_2$ norm to improve the performance.

To include statistical data information in TWSVM, Peng and Xu \cite{peng2012twin} formulated twin Mahalanobis distance-based SVM (TMSVM) that uses each classes covariance to determine hyperplanes. It has better training speed and generalization than TWSVM. Shao et al. \cite{shao2012probabilistic}  in 2012 introduced a probability based TWSVM model (P-TWSVM) which  is more accurate than TWSVM. In 2012, Shao and Deng \cite{shao2012coordinate} formulated a coordinate descent based TWSVM to increase the efficiency of TWSVM by introducing a regularization term. Further, to solve the dual problems, the authors proposed a coordinate descent method which reduces the training time even in case of large datasets as it deals with single data point at once.

To include the underlying correlation information between data points in TWSVM, in 2012, Ye et al. \cite{ye2012weighted} formulated a novel nonparallel plane classifier, called Weighted TWSVM with Local Information (WL-TWSVM). A major limitation of this algorithm is that it doesn't work for large-scale problems as it  finds the $K$-nearest neighbors for all the samples. Based on TWSVM, in 2013, Peng and Xu \cite{peng2013twin} formulated a twin-hypersphere SVM (TH-SVM). Unlike TWSVM, it determines two hyperspheres and avoids the matrix inversions in its dual formulations.

To incorporate the structural data information in TWSVM, Qi et al. \cite{qi2013structural} formulated structural-TWSVM which is superior in training time as well as accuracy to that of TWSVM. However, it still ignores the importance of different samples within each cluster. To overcome this drawback, in 2015, Pan et al. \cite{pan2015k} uses $K$-nearest neighbor based structural TWSVM which provides different penalties to the samples of different classes. However, it suffers from overfitting due to empirical risk minimisation. To reduce the overfitting issues, efficient KNN weighted TWSVM \cite{xie2019efficient} introduced the regularisation term to avoid the issues of overfitting. This also ensured the minimization of structural risk which results in better generalization. 

In 2014, Shao et al. \cite{shao2014nonparallel} formulated a different nonparallel hyperplane SVM in which hyperplanes are determined by clustering the samples based on similarity between the two classes. This algorithm has better or comparable accuracy with low computational time. To optimally select the parameters in TWSVM, in 2016 Ding et al. \cite{ding2016twin} proposed TWSVM build on Fruit Fly Optimization Algorithm (FOA-TWSVM). It can optimally select the parameters in less time and with better accuracy. In 2017, Pan et al. \cite{pan2017safe} proposed safe screening rules to make TWSVM efficient for large-scale classification problems. 
The safe screening rules reduce the scale by eliminating training samples and giving same solution as the original problem. Experimental results show that safe screening rules can greatly reduce the computational time while giving the same solutions as original ones. Yang and Xu \cite{yang2018safe} proposed safe sample screening rule (SSSR) for Laplacian twin parametric-margin SVM (LTPSVM) to address the problems while handling large-scale problems. In 2018, Pang and Xu \cite{pang2019safe} proposed safe screening rule for Weighted TWSVM with local information (WLTWSVM) to implement the algorithm for large-scale datasets. Experimental results demonstrate the effectiveness of SSSR for WLTWSVM as it performes better than SVM and TWSVM with significantly less computational time. Recently, Zhao et al. \cite{zhao2019improved} proposed an efficient non-parallel hyperplane Universum support vector machine (U-NHSVM) for classification problems. U-NHSVM is flexible to exploit the prior information in universum.

Tanveer \cite{tanveer2015application} proposed unconstrained minimization problem (UMP) formulation of Linear programming TWSVM to enhance robustness and sparsity in TWSVM. Tanveer \cite{tanveer2015newton} also proposed an implicit Lagrangian TWSVM which is solved by using finite Newton method. Tanveer and Shubham \cite{tanveer2017smooth} in 2017 proposed smooth TWSVM via UMP which increases the generalization ability and training speed of TWSVM.

Multi-label learning deals with data having multiple labels and has gained a lot of attention recently. Chen et al. \cite{chen2016mltsvm} in 2016 proposed multi-label TWSVM (MLTWSVM) that exploits multi-label information from instances. In 2020, Azad\textendash Manjiri et al. 
 \cite{azad2020ml} proposed structural twin support
vector machine for multi-label learning (ML-STWSVM) which embeds the prior structural information of data into the optimization function of MLTWSVM based
on the same clustering technology of S-TWSVM. This algorithm achieved better performance compared to other baseline multi-label learning algorithms. 

In 2018, Rastogi et al. \cite{rastogi2018generalized} proposed a new loss function termed as $(\epsilon_1, \epsilon_2)$-insensitive zone pinball loss function which generalizes other existing loss functions e.g. pinball loss, hinge loss. The resulting model takes care of noise insensitivity, instability of re-sampling and scatterdness present in the datasets. In 2019, Tanveer et. al \cite{tanveer2019comprehensive} presented rigorous comparison of 187 classifiers which includes 8 variants of TWSVM and exhaustive evaluation of these classifiers was performed on 90 UCI benchmark datasets. Results have shown that RELS-TSVM achieved highest performance than all other classifiers for binary classification task. Thus,  RELS-TSVM is the best TWSVM classifier. For more details, one can refer to \cite{tanveer2019comprehensive}. In 2020, Ganaie and Tanveer \cite{ganaie2020lstsvm} proposed a novel classification approach using pre-trained functional link to enhance the feature space. Authors performed the classification task by LSTWSVM on the enhanced features and validated the performance on various datasets. Some other recent research on TWSVM include \cite{ganaie2020oblique} where authors proposed a novel way for generating oblique decision trees. The classification of training samples is done based on the Bhattachrayya distance with randomly selected feature subset and then hyperplanes are generated using TBSVM. The major advantage of the proposed model is that there is no need for any extra regularization as matrices are positive definite. The ensemble \cite{ganaie2021ensemble} based models of twin SVM based models was proposed in \cite{tanveer2021ensemble}.

\subsection{Twin Support Vector Machine for Multi-class Classification}

Earlier, TWSVM was only implemented to solve binary class problems, however, the majority of problems in the real-world applications are generally based on multicategory classification. Thus, Xu et al. \cite{xu2013twin} formulated Twin-KSVC. It implements ``1-versus-1-rest" form to provide ternary output $(-1, 0, 1)$. This algorithm requires to solve two smaller-sized QPPs. It has better accuracy than 1-versus-rest TWSVM but losses sparsity. Yang et al. \cite{yang2013multiple} in 2013, proposed multiple birth SVM (MBSVM). It is computationally better than TWSVM even when number of classes are large.

In 2013, Xie et al. \cite{xie2013extending}  extended TWSVM application for multi-class problems and proposed one-versus-all TWSVM (OVA-TWSVM) which solve $k$-category problem using one-versus-all (OVA) approach to develop $k$ TWSVM. To strengthen the performance of multi-TWSVM, Shao et al. \cite{shao2013best} formulated a separating decision tree TWSVM (DTTWSVM). The basic idea of DTTWSVM is to embody the best separating principle rule to create a binary tree and then built binary TWSVM model on each node. Experimental results have shown that DTTWSVM has low computational complexity and better generalization.

Xu and Guo \cite{xu2014twin} in 2014 formulated twin hyper-sphere multi-class SVM (THKSVM) which employs the ``rest-versus-1'' structure instead of  ``1-versus-rest'' structure in TWSVM. In order to find hyperspheres, it constructs $k$ (no. of classes) classifiers and finds $k$ centers and $k$ radiuses for each hypersphere. Each hypersphere covers maximum points of $K_1$ classes and is as distant as possible from the rest classes. It has fast computational speed as compared to Twin-KSVC and also inverse operation of matrices are not required while solving dual QPPs of THKSVM. Thus, it performs better on large datasets and has better accuracy than``1-versus-rest'' TWSVM but lower than Twin-KSVC. To enhance the performance of Twin-KSVC, in 2015, Nasiri et al. \cite{nasiri2015least} formulated LS version of Twin-KSVC that works similar to Twin-KSVC but  solves linear system of equations rather than pair of QPPs in Twin-KSVC. The proposed algorithm is fast, simple and has better accuracy and lower training time than Twin-KSVC. Another approach to enhance the performance of Twin-KSVC and to include local information of samples, in 2016, Xu \cite{xu2016k} proposed $K$-nearest neighbor-based weighted multi-class TWSVM which uses information from within class and applies a weight in the objective function. This algorithm has low computational cost and better accuracy than Twin-KSVC.

Based on the multi-class extension of the binary LS-TWSVM,   weighted multi-class LSTWSVM algorithm \cite{tomar2015effective} and regularized least squares twin SVM  \cite{ali2022regularized} have been proposed for multi-class imbalanced data. To control sensitivity of classifier, weight setting is employed in loss function for determining each hyperplane. Experimental results have shown the superiority and feasibility of the proposed algorithm for multi-class imbalanced problems. The authors \cite{tomar2015comparison} extended LS-TWSVM for multi-class classification and compared various concepts of a multi-class classifier like ``One-versus-All", ``One-versus-One", ``All-versus-One" and ``Directed Acyclic Graph (DAG)". DAG MLS-TWSVM performance is superior and has high computational efficiency.

In 2016, Yang et al. \cite{yang2016least} formulated least squares recursive projection TWSVM. For each class, it determines $k$  projection axes and needs to solve linear equations. It has similar performance as MP-TWSVM. Based on P-TWSVM, Li et al. \cite{li2016multiple} in 2016, proposed multiple recursive projection TWSVM (Multi-P-TWSVM) which solves $k$ QPPs in order to determine $k$-projection axes (for $k$ classes). Authors introduced regularization term and recursive procedure which increases the generalization but this algorithm is complex when more orthogonal projection axes are generated.

In 2017, Ding et al. \cite{ding2017review} review various multi-class algorithms as per their structures: ``one-versus-rest'', ``one-versus-one'',  ``binary tree'', ``one-versus-one-versus-rest'', ``all-versus-one''  and ``direct acyclic graph'' based multi-class TWSVM. All these multi-class TWSVMs have some advantages and disadvantages. In general, one-versus-one TWSVMs have higher performance. Ai et al. \cite{ai2018multi} in 2018 proposed a multi-class classification weighted least squares TSVH using local density information in order to improve the performance of LSTSVH. Authors introduced local density information into LSTSVH to provide weight for each data point in order to avoid noise sensitivity. In 2018, Pang et al. \cite{pang2018scaling} proposed $K$-nearest neighbor-based weighted multi-class TWSVM (KMTWSVM) that incorporated "1-versus-1-versus-rest" strategy for multi-class classification and also takes into account the distribution of all instances. However, it is computationally extensive especially for large-scale problems. Thus, authors in \cite{pang2018scaling} also proposed safe instance reduction rule (SIR-KMTWSVM) to reduce its computational time. Lima et al. \cite{de2018improvements} proposed improvements on least squares
twin multi-class classification SVM which is ``one-versus-one-versus-rest'' strategy and generated ternary output. The proposed algorithm only needs to deal with linear equations. Numerical results demonstrate that it achieves better classification accuracy than Twin-KSVC \cite{xu2013twin} and LSTKSVC \cite{nasiri2015least}. 
Qiang et al. \cite{qiang2020robust}  proposed improvement on LSTKSVC by proposing robust weighted linear loss twin multi-class support vector machine (WLT-KSVC) which takes care of the two drawbacks of LSTKSVC;  sensitive to outliers and misclassifying some rest class samples due to the use of quadratic loss. Experiments on the UCI and NDC datasets showed promising results of this algorithm but its training accuracy significantly decreases as the number of classes increases. Li et al. \cite{li2019single} proposed a nonparallel support vector machine (NSVM) for multiclass classification problem. Numerical experiments on several benchmark datasets clearly show the advantage of NSVM. In 2020, Tanveer et al. \cite{tanveer2020least} proposed a fast and improved version of KWMTWSVM \cite{xu2014k} called least square K-nearest neighbor weighted multi-class TWSVM (LS-KWMTWSVM). Numerical experiments on various KEEL imbalance datasets showed high accuracy and low computational time for the proposed LS-KWMTWSVM as compared to other baseline algorithms. A nulticlass nonparallel parametric margin SVM \cite{du2021multiclass} has also been proposed for multiclass classification. Kernel free least squares TWSVM \cite{gao2021novel} via special fourth order polynomial surface  resulted in improved performance in multiclass problems.

\subsection{Twin Support Vector Machine for Semi-Supervised Learning}

Semi-supervised learning (SSL) techniques have achieved extensive attention from many researchers in the last few years due to its promising applications in machine learning and data mining. In many real-world challenges, labeled data is not easily available and thus it deteriorates the performance of supervised learning algorithms due to insufficient supervised information. SSL overcomes this limitation and uses both unlabeled and labeled data.

In 2012, Qi et al. \cite{qi2012laplacian} formulated a novel Laplacian \textendash TWSVM for the semi-supervised classification problem. This algorithm assumes that data points lie in low dimensional space and uses the geometric information of the unlabeled data points. Similar to TWSVM it solves a pair of QPPs with inverse matrix operations. Results have shown that Lap-TWSVM has better flexibility, prediction accuracy and generalization performance than conventional TWSVM. This algorithm has excellent performance for semi-supervised classification problems but due to QPPs and inverse matrix operations, its computational cost is high. Thus, to increase the training speed of Lap-TWSVM, Chen et al. \cite{chen2014laplacian} in 2014 formulated LS version of Lap-TWSVM (Lap-LS-TWSVM). Unlike Lap-TWSVM, it needs to deal with linear equations and is done using conjugate gradient algorithm. It has less training time than Lap-TWSVM. Another algorithm was formulated by Chen et al. \cite{chen2014laplacian1} in 2014 to improve the performance of Lap-TWSVM by transforming the QPPs to UMPs in primal space and smoothing method is used which is effectively solved by Newton-Armijo algorithm. It achieved comparable accuracy as Lap-TWSVM with less computational time. In 2016, Khemchandani and Pal \cite{khemchandani2016multi} extended this to multi-class classification and formulated Lap-LS-TWSVM which evaluates training samples to ``1-versus-1-versus-rest'' and provides ternary output $(-1, 0,+1)$. It has better accuracy and less training time than Lap-TWSVM.

Based on the similar idea of Lap-TWSVM, in 2016, Yang and Xu \cite{yang2016laplacian} proposed an extension of traditional TPMSVM, which makes use of graph Laplacian and creates a connection graph of the training data points whose solution can be obtained by solving two SVM-type QPPs. Experiments reveal that LTPMSVM has higher classification accuracy than SVM, TPMSVM, Lap-TWSVM, and TWSVM.

Khemchandani and Pal \cite{khemchandani2017tree}, in 2017, blended the Laplacian\textendash TWSVM and Decision Tree \textendash TWSVM classifier and formulated a tree based classifier for semi-supervised multi-class classification. Extensive experiments on color images shows the feasibility of the model. Another interesting approach in this direction in 2019, has been suggested by Rastogi and Sharma  \cite{rastogi2019fast} called as Fast Laplacian TWSVM for Active Learning ($FLap-TWSVM_{AL}$) where the authors proposed to identify the most informative and representative training points whose labels are queried for domain experts for annotations. Once the corresponding labels are acquired, this limited labeled and unlabeled data are used to train a fast model that involves solving a QPP and an unconstrained minimization problem to seek the classification hyperplanes.

\subsection{Twin Support Vector Machine for Clustering}

Wang et al. \cite{wang2015twin} formulated twin support vector clustering algorithm (TSVC) build upon TWSVM. It divides the data samples into $k$ clusters such that the data samples are around $k$ cluster center points. It exploits the information from clusters (both between and within clusters) and the center planes are determined by solving a series of QPP. Authors in  \cite{wang2015twin} also proposed a nearest neighbor graph (NNG)-based initialization to make the model more stable and efficient. Improved TSVC \cite{moezzi2019twsvc+} decomposed the multiclass clustering problem into multiple two cluster problems.

In 2018, Khemchandani et al. \cite{khemchandani2018fuzzy} formulated fuzzy least squares version of TSVC in which each data point has a fuzzy membership value and is allotted to different clusters. The proposed algorithm solves primal problems instead of dual problems in TSVC. Experimental results shown that the proposed algorithm obtains comparable clustering accuracy to that of TSVC but has less training time.

Khemchandani and Pal \cite{khemchandani2016weighted} replaced the hinge loss function of TSVC with weighted linear loss and introduced a regularization term in TSVC which needs to solve linear equations and also implements the structural risk component. The proposed algorithm called weighted linear loss TSVC achieves higher accuracy than TSVC. The loss function proposed in \cite{khemchandani2016weighted} is not continuous and therefore, authors in \cite{rastogi2019fuzzy} modified the weighted linear loss function to form a continuous loss function and implemented TSVC and WLL-TSVC in semi-supervised framework and also introduced a fuzzy extension of semi-supervised WLL-TSVC to make a robust algorithm which is less sensitive to
noise. Experimental results have demonstrated that the proposed algorithm achieved better clustering accuracy and less computational time than TSVC and WLL-TSVC. Tree-based localized fuzzy twin support vector clustering (Tree-TSVC) was proposed by \cite{poojaclustering}. Tree-TSVC is a novel clustering algorithm that builds the cluster that represents a node on a binary tree, where each node comprises of proposed TWSVM based classifier. Due to the tree structure and the formulation that leads to solving a series of systems of linear equations, Tree-TSVC model is efficient in terms of training time and accuracy. 

TSVC uses squared $L_2$-norm distance which leads to noise sensitivity. Also, each cluster plane learning iteration requires to solve a QPP which makes it computationally complex. To address these issues in TSVC, Ye et al. \cite{ye2018l1} in 2018 used $L_1$ norm distance and proposed $L_1$-norm distance minimization-based robust TSVC which only deals with linear equations instead of series of QPPs. Authors in \cite{ye2018l1} further proposed RTSVC and Fast RTSVC to speed up the computation of TSVC in nonlinear case. Numerical experiments shown that this model has higher accuracy as compared to other k-clustering methods and has less computational time than TSVC. Wang et al. \cite{rampwang2018} proposed ramp-based TSVC by introducing the ramp cost function in TSVC. The solution of the proposed algorithm is obtained by solving non-convex programming problem using an alternating iteration algorithm. In 2019, Bai et al. \cite{bai2019clustering} introduced regularization in clustering and proposed large margin TSVC. Authors in \cite{bai2019clustering}  proposed a fast least squares TSVC with uniform coding output. The algorithm achieves better performance than TSVC and other plane-clustering methods. Wang et al. \cite{wang2019general} in 2019 proposed a general model for plane-based clustering which includes various extensions of k-plane clustering (kPC). It optimizes the problem by minimizing the total loss which is derived from both within-cluster and between-cluster. It can capture data distribution accurately. In 2020, Richhariya and Tanveer \cite{richhariya2020least} proposed projection based least square TSVC (LSPTSVC) and employed the concave-convex procedure (CCCP) to solve the optimization problem. LSPTSVC only needs to solve a set of linear equations and thus leads to significantly less computational time. TSVC uses hinge loss function, hence suffers from the issues of noise and has low sampling stability. To overcome these issues, pinball loss twin support vector clustering \cite{tanveer2021pinball} used pinball loss to penalize the samples. However, it suffers from the issues of overfitting as it minimises the empirical risk. Hence, pinball loss twin bounded support vector clustering \cite{tanveer2021pinballcon} introduced the regularisation term to minimise the structural risk and avoids the issues of overfitting. 
Introduction of pinball loss function leads to loss of sparsity in the model. Hence, sparse pinball loss TSVC \cite{tanveer2021sparse,tanveertabish2021} used $\epsilon$-insensitive pinball loss function  to make the model sparse. It implements the empirical risk minimisation principle and thus suffers due to overfitting. To minimise the overfitting issues, sparse pinball twin bounded support vector clustering \cite{tanveer2021pinballbo} used the regularisation term to minimise the structural risk and hence, avoid the issues of overfitting.  

\section{Applications of Twin Support Vector Classification}
\label{sec:Applications of Twin Support Vector Classification}
\begin{figure}
    \centering
    \includegraphics[width=\linewidth]{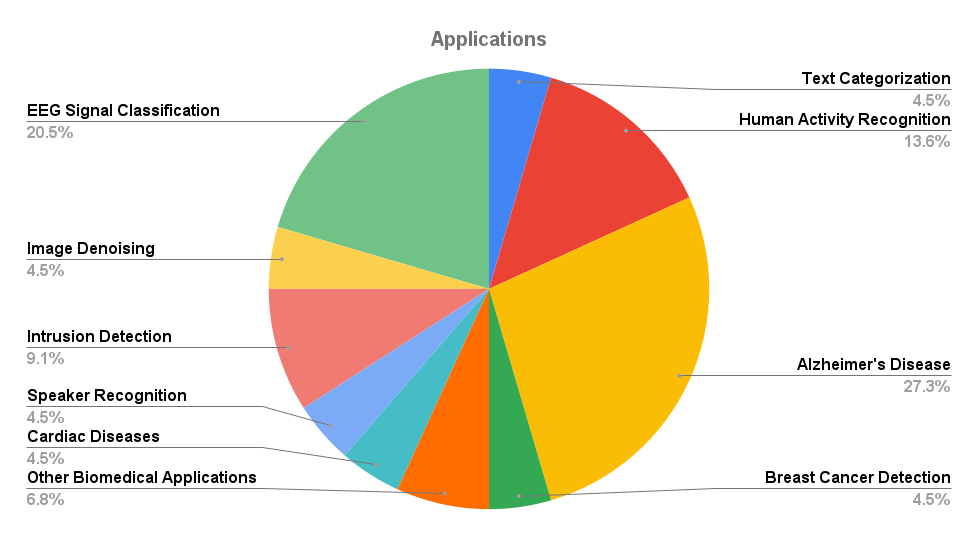}
    \caption{Applications of Twin Support Vector Classification}
    \label{fig:Applications of Twin Support Vector Classification}
\end{figure}
TWSVM has been implemented to solve many real-life classification challenges and has shown promising results in some  applications. 
%TWSVM being a classifier has been applied for many pattern classification problems and researchers have proposed many improved algorithms and variants for pattern classification

\subsection{Biomedical applications}
TWSVM has been applied to detect breast cancer in early stages by detecting a mass in digital mammograms \cite{si2009mass}. A hybrid feature selection based approach has also been implemented for detecting breast cancer, Hepatitis, and Diabetes \cite{tomar2015hybrid}. 
Wang et al. \cite{wang2016dual} proposed TWSVM in combination with dual-tree complex wavelet transform (DTCWT) method for Pathological Brain Detection. 
TWSVM is also implemented for detecting cardiac diseases, one such application was proposed by Houssein et al. in \cite{houssein2018improving}, heartbeats were detected using Swarm-TWSVM and this algorithm achieved better accuracy than TWSVM. Refahi et al. \cite{refahi2018ecg} used LSTWSVM and DAG LS-TWSVM classifiers for predicting arrhythmia heart disease. Chandra and Bedi \cite{chandra2018twin} proposed linear norm fuzzy based TWSVM for color based classification of human skin which achieved better accuracy than other conventional methods. Xu et al. \cite{xu2015imbalanced} proposed semi-supervised TWSVM for detection of Acute-on-chronic liver failure (ACLF).
Also, many researchers have applied TWSVM to detect Alzheimer's disease in its early stages \cite{tanveer2019ADreview,zhang2015detection,wang2016detection,alam2017twin,tomar2014feature,wang2016dual,tomar2014efficient,wang2016morphological,tomar2014emotion,wang2015pathological}.

\subsection{Alzheimer's Disease Prediction}
In 2020, Richhariya and Tanveer \cite{richhariyaefficient} proposed an angle based universum least squares TWSVM (AULSTWSVM) which performed with 95\% accuracy for detecting Alzheimer's disease (AD). Authors \cite{richhariya2020diagnosis} also proposed universum support vector machine based recursive feature elimination (USVM-RFE) for detecting AD. Khan et al. \cite{khannovel} proposed an approach to improve the classification accuracy in mild cognitive impairment (MCI), normal control (NC), and AD subjects using structural magnetic resonance imaging (sMRI). Authors used FreeSurfer to process MRI data and derive cortical features which are used in TWSVM, LSTWSVM and RELS-TSVM to detect AD. Sharma et al. \cite{sharma2022fdn} proposed fuzzy LS-TWSVM based deep learning network for prognosis of the Alzheimer’s disease.

\subsection{Speaker Recognition}
Cong et al. \cite{cong2008efficient} formulated multi-class TWSVM for speaker recognition with feature extraction based on gaussian mixture models (GMMs). It gives better results than traditional SVM. Zhendong and Yang \cite{yang2009study} proposed multi-TWSVM which find hyperplane for every class and takes constraints from other classes separately on the QPP. This algorithm performed better than many other algorithms for speaker recognition.

\subsection{Text Categorization}
Kumar and Gopal \cite{kumar2009least} formulated a least squares TWSVM and experiments have shown the validity of the model for text applications. Francis and Sreenath \cite{francis2019robust} proposed manifold regularized TWSVM for text recognition and the proposed method achieved highest accuracy among SVM, LSTWSVM and other methods. Non-parallel SVM \cite{tian2014nonparallel} and efficient pinball TWSVM \cite{rastogi2021efficient} has also demonstrated better performance in text categorization.

\subsection{Intrusion Detection}
It is a system which monitors or protects the network against any malicious activity. It has been a critical component for network security. Researchers \cite{he2014intrusion,ding2008high,mousavi2015semi,nie2013application} applied TWSVM for intrusion detection and results have shown that it achieves better accuracy than other intrusion detection algorithms. 

\subsection{Human activity recognition}
Khemchandani and Sharma \cite{khemchandani2016robust} proposed least square TWSVM for human activity recognition which gives promising results even with the outliers. Nasiri et al. \cite{nasiri2014energy} formulated energy-based LS-TWSVM algorithm. Khemchandani and Sharma \cite{khemchandani2017robust} also proposed robust parametric twin support vector machine which can effectively deal with the noise. Mozafari et al. \cite{mozafari2011action} used the Harris detector algorithm and applied LS-TWSVM for action recognition and achieved the highest accuracy than other state-of-the-art methods. Kumar and Rajagopal \cite{kumar2018detecting} proposed Multi-class TWSVM for detecting human face happiness combined with Constrained Local Model.
Authors \cite{kumar2019detecting} also proposed semi-supervised multi TWSVM to predict human facial emotions with $13$ minimal features that can detect six basic human emotions. Algorithm achieved highest accuracy and least
computation time with minimal feature vectors.

\subsection{Image Denoising}
Yang et al. \cite{yang2014image} proposed edge/texture-preserving image denoising based on TWSVM which is very effective to preserve the informative features such as edges and textures and better than other image denoising methods available. Shahdoosti and Hazavei \cite{shahdoosti2018combined} proposed a ripplet formulation of the total variation method for denoising images. This algorithm attains promising results in improving the image quality in terms of both subjective and objective inspections.

\subsection{Electroencephalogram (EEG)}
Classification of electroencephalogram (EEG) for different mental activities has been an active research topic. Richhariya and Tanveer \cite{richhariya2018eeg} proposed universum TWSVM which is insensitive to outliers as it selects universum from the EEG datasets itself to generate universum data points which remove the effect of outliers. Soman and Jayadeva \cite{soman2015high} used the classifiable metric to choose discriminative frequency bands and used the TWSVM to learn imbalanced datasets. Tanveer et al. \cite{tanveer2018entropy} proposed entropy based features in Flexible analytic wavelet transform (FAWT) framework and RELS-TSVM \cite{tanveer2016robust}  for classification to detect epileptic seizure EEG. Tanveer et al.  \cite{tanveer2018classification} used FAWT framework for classification of seizure and seizure-free EEG signals with Hjorth parameters as features and implemented TWSVM, LSTWSVM and RELS-TSVM \cite{tanveer2016robust} for classifying signals. Li et al. \cite{li2018self} proposed LSTWSVM with a frequency band selection common spatial pattern algorithm for detecting motor imagery electroencephalography. This algorithm achieved faster training time compared to other SVM baseline algorithms.
Some more researchers have applied TWSVM for EEG classification, biometric identification and other leak detection challenges \cite{kumar2021universum,gupta2021data,she2015multiclass,li2018self,kostilek2012eeg,lang2017leak,dalal2019automated}

\subsection{Other Applications} 
Cao et al. \cite{cao2018improved} proposed improved TWSVM with multi-objective cuckoo search to predict software bugs. Authors employed TWSVM to predict defected modules and used cuckoo search to optimize TWSVM and this proposed method achieved better accuracy than other software defect prediction methods. Chu et al. \cite{chu2018multi} proposed Multi-information TWSVM for detecting steel surface defects. The TWSVM models have also been used in  image recognition or face recognition \cite{qi2013robust,peng2012twin,chen2018new,peng2013bi} and facial expression recognition \cite{richhariya2019facial} and privacy preservation \cite{anand2019privacy}.
\newline 
\newline
\setlength{\tabcolsep}{6pt} % Default value: 6pt
\renewcommand{\arraystretch}{2} % Default value: 1
\hspace*{-2.5cm}
% \begin{small}
\begin{table}%[h]
\caption {Optimization framework for TWSVM algorithms}
\label{tab:Optimization framework}
\resizebox{\textwidth}{!}{%
\begin{tabular}{p{2.2cm} p{1.8cm}p{2.4cm}p{2.5cm}p{1.1cm}p{1.3cm}}
% {c c c c c c }

\hline
Method & Regularization term & $\underline D(g_y(X_y),0 )$ & $\overline{D}(g_y(X_j),g_y(X_y))$ & $D(g_y(x))$ & Distance\\
\hline
GEPSVM \cite{mangasarian2006multisurface} & $\norm{\begin{bmatrix}
w_{y}\\b_{y}\end{bmatrix}}^2$ &$\norm{g_y(X_y)}^2$ & $\norm{g_y(X_j)}^2$ & $\frac{\abs{g_y(x)}}{\norm{w_y}}$ & $\norm{\cdot}$ \\
\hline
TWSVM \cite{khemchandani2007twin}&$0$ & $\norm{g_y(X_y)}^2$  & $e^T\norm{g_y(X_j)-e}_+$ & $\abs{g_y(x)}$ & $\norm{\cdot}/(.)_+$  \\
\hline
TBSVM \cite{shao2011improvements} &$\norm{\begin{bmatrix}
w_{y}\\b_{y}\end{bmatrix}}^2$ &$\norm{g_y(X_y)}^2$ &$e^T\norm{g_y(X_j)-e}_+$ &$\frac{\abs{g_y(x)}}{\norm{w_y}}$
&$\norm{\cdot}/(.)_+$\\
\hline
LSTWSVM \cite{kumar2009least} &$0$ &$\norm{g_y(X_y)}^2$ & $\norm{g_y(X_j)-e}^2$ &$\abs{g_y(x)}$ & $\norm{\cdot}$\\
\hline
Pin-TWSVM \cite{xu2016novel} & $\norm{w_y}^2$ & $L_\tau(X_y,y,g_y,(X_y))$ & ${g_y(X_j)}_1$ & $\frac{\abs{g_y(x)}}{\norm{w_y}}$ & $\norm{\cdot}/L_T(.)/\norm{\cdot}_1$ \\
\hline
Pin-GTSVM \cite{tanveer2019general} & $\norm{w_y}^2$ & $\norm{g_y(X_y)}^2$ & $L_\tau(X_y,y,g_y,(X_y))$ & $\frac{\abs{g_y(x)}}{\norm{w_y}}$& $\norm{\cdot}/L_T(.)/\norm{\cdot}_1$\\
\hline
NPSVM \cite{tian2014efficient}& $\norm{w_y}^2$ & $\norm{g_y(X_y)-\epsilon}_1$ & ${e^T(g_y(X_j)-e)}_+$ & $\abs{g_y(x)}$ & $\norm{\cdot}/\norm{\cdot}_1/(.)_+$\\
\hline

Par-TWSVM \cite{peng2011tpmsvm} & $\norm{w_y}^2$ &$ e^Tg_y( X_j )$ & $\norm{g_y( X_y)}_{+}$ & $\abs{g_y( x )}$ & $\norm{\cdot}$\\
\hline
\end{tabular}}
\end{table}
% \end{small}

The pinball loss function is defined as follows:
\begin{align}
L_\tau\left(X_y,y,g_y,\left(X_y\right)\right)
= \left\{\begin{array}{cc}
e^T(0-yg_y(X_y)),& (0-yg_y(X_y))\ge 0, \\
-\tau e^T(yg_y(X_y)-0),& (0-yg_y(X_y))< 0.
\end{array}\right.
\end{align}

where $g_y(x)=g(x;w_y,b_y)=w_y^Tx+b_y=0$,  $\underline{D}(g_y(A,0))$ denotes the intraclass distance  which represents objective function while
$\overline{D}(g_y(A),g_y(B))$ is interclass distance which corresponds to constraints, $D(g_y(x))$ is the perpendicular distance of point $x$ from the hyperplane $g_y(x)=0$. 

Optimization framework of various TWSVM algorithms are discussed in Table \ref{tab:Optimization framework} \cite {li2019single}.

Table \ref{tab:Properties of different} shows the differences in major TWSVM methods based on the SRM principle, sparsity, matrix inversion and noise insensitivity.

\hspace*{-2.5cm}
% \begin{small}
\begin{table}%[h]
\caption {Properties of different TWSVM Algorithms} \label{tab:Properties of different}
\resizebox{\textwidth}{!}{%
\begin{tabular}{l c c c c }

\hline
Models $\backslash$ Characteristics & SRM & Sparsity & Matrix Inversion & Insensitive to noise  \\\hline
 TWSVM \cite{khemchandani2007twin}  &  & \checkmark & \checkmark & \checkmark\\
 TBSVM \cite{shao2011improvements}  & \checkmark & \checkmark & \checkmark & \checkmark\\
 LSTWSVM \cite{kumar2009least}  &  &   & \checkmark & \\ 
 RELS-TSVM \cite{tanveer2016robust} & \checkmark  &   & \checkmark & \checkmark\\
Projection TWSVM  \cite{hua2017novel} & \checkmark  & & \checkmark & \checkmark\\
TPMSVM \cite{peng2011tpmsvm} &\checkmark  &\checkmark & & \checkmark\\
Pin-GTSVM \cite{tanveer2019general} &  &  & \checkmark  & \checkmark \\ 
SP-TWSVM \cite{tanveer2019sparse} &  & \checkmark & \checkmark  &  \checkmark\\
Pin-TPMSVM \cite{xu2016novel} & \checkmark &  &  &\checkmark  \\ 
ISPTWSVM \cite{tanveer2019improved} & \checkmark & \checkmark & \checkmark &\checkmark  \\
\hline
\end{tabular}
}
\end{table}

\section{Basic theory of Twin Support Vector Regression}
\label{sec:Basic theory of Twin Support Vector Regression}

% In 2010, Peng \cite{peng2010tsvr} introduced a new nonparallel plane regression for the first time, termed as the twin support vector regression (TSVR). Based on TWSVM, TSVR also aims at generating two nonparallel functions such that each function determines the ɛ-insensitive down or up- bounds of the unknown regressor. TSVR formulation is different that SVR in two ways, TSVR requires to solve two smaller QPPs instead of one large QPP in SVR and TSVR has half the group of constraint required for all the data points than SVR.

% \subsection{Twin support vector regression (TSVR)}

Peng \cite{peng2010tsvr} in 2010 proposed an efficient twin support vector regression (TSVR) algorithm in line with TWSVM, called twin support vector regresson (TSVR). Like TWSVM, it also requires to solve two QPPs. It finds an end regressor that is the mean of $\varepsilon$-insensitive up and bound functions. TSVR has less computational time than a standard SVR and has better generalization ability. The down- and up-bound functions for linear case is given below:

For any $x\in R^{n}$, the two hyperplanes are defined as follows:

\begin{align}
\label{eqn:9}
f_1 (x)=u_{1}^{T} x+b_{1}  \quad \mbox{and} \quad f_{2} (x)=u_{2}^{T} x+b_{2},
\end{align}       
\noindent
% respectively, such that the unknowns $w_{1} ,w_{2} \in R^{n} $ and $b_{1} ,b_{2} \in R$ become the solutions of the following pair of QPPs \cite{peng2010tsvr}:
The two QPPs in linear case are defined as below:
\begin{align}
& \mathop{\min }\limits_{(u_{1} ,b_{1} ,\zeta_1  )\in R^{n+1+m} } \frac{1}{2} \norm{y- e\varepsilon _{1} -(Au_{1} +b_{1} e)}^{2} +C_{1} e_{}^{T} \zeta_1   \notag\\
&  \, \quad\quad\quad s.t. \,\quad\quad \quad  y-(Au_{1} +b_{1} e)\ge e \varepsilon _{1} -\zeta_1, \quad \zeta_1  \ge 0
\end{align}
and
\begin{align}
& \mathop{\min }\limits_{(u_{2} ,b_{2} ,\zeta_2 )\in R^{n+1+m} } \frac{1}{2} \norm{y+e \varepsilon _{2} -(Au_{2} +b_{2} e)}^{2} +C_{2} e_{}^{T} \zeta_2 \notag \\
& \,\, \quad\quad  \quad s.t. \,\quad \quad \quad  (Au_{2} +b_{2} e)-y\ge  e \varepsilon _{2} -\zeta_2,~ \zeta_2 \ge 0,         
\end{align} 
 here, $C_{1} ,C_{2} >0$; $\varepsilon _{1} ,\varepsilon _{2} >0$ and $\zeta_1 ,\zeta_2 $ are slack variables.
The final regressor is the mean of up and down regressors in (\ref{eqn:9}), which is given as follows
\begin{align}
f(x)=\frac{1}{2} (f_{1} (x)+f_{2} (x)) \quad \mbox{for all} \quad x\in R^{n}.
\end{align}

For nonlinear case, kernel surfaces are used rather than hyperplanes which are given below:

% For the input matrix $A\in R^{m\times n} $ and a nonlinear kernel function $k(.,.)$ given,  let the kernel matrix $k(A,A^{t} )$ of order\textit{ m }whose (i,j)-th  element be defined by: $k(A,A^{t} )_{ij} =k(x_{i} ,x_{j} )\in R$ and let  $k(x^{t} ,A^{t} )=(k(x,x_{1} ),...,k(x,x_{m} ))$ be a row vector in  $R^{m} $.

% The nonlinear TSVR seeks the down- and up- bounds of the form:
\begin{align}
f_{1} (x)=k(x^{T} ,A^{T} )u_{1} +b_{1} \quad \mbox{and} \quad f_{2} (x)=k(x^{T} ,A^{T} )u_{2} +b_{2}.
\end{align}

The two QPPs in non-linear case are defined as below:
\begin{align}
& \mathop{\min }\limits_{(u_{1} ,b_{1} ,\zeta_1 )\in R^{m+1+m} } \frac{1}{2} \norm{y-e \varepsilon _{1} -(k(A,A^{T} )u_{1} +b_{1} e)}^{2} +C_{1} e_{}^{T} \zeta_1  \notag\\
&\quad \quad \quad s.t. \quad\quad\quad y-(k(A,A^{T} )u_{1} +b_{1} e)\ge e\varepsilon _{1} - \zeta_1  ^{} , \zeta_1  \ge 0
\end{align}
and
\begin{align}
&\mathop{\min }\limits_{(u_{2} ,b_{2} ,\zeta_2 )\in R^{m+1+m} } \frac{1}{2} \norm{y+e \varepsilon _{2} -(k(A,A^{t} )u_{2} +b_{2} e)}^{2} +C_{2} e_{}^T \zeta_2 \notag\\ 
& \,\quad\quad \quad s.t. \quad\quad\quad  (k(A,A^{T} )u_{2} +b_{2} e)-y\ge e \varepsilon _{2} -\zeta_2 ^{} , \zeta_2 \ge 0.
\end{align}
For more details, one can refer to \cite{peng2010tsvr}. 

Although taking motivation from TWSVM formulation, Peng  \cite{peng2010tsvr} attempted to propose TSVR where the regressor is obtained via solving a pair of quadratic programming problems (QPPs). However, later authors in \cite{khemchandani2016twsvr}  argued that TSVR formulation is not in the true spirit of TWSVM. Further, taking the motivation from Bi and Bennett \cite{Biandbennett}, they proposed an alternative approach to find a formulation for 
TSVR which is in the true spirit of TWSVM. They have shown that their proposed TSVR  formulation can be derived from TWSVM for an appropriately constructed classification problem.

\section{Research progress on Twin Support Vector Regression}
\label{sec:Research progress on Twin Support Vector Regression}
In this section, we discuss the progress of twin SVM based models for the regression problems.

\subsection{Weighted Twin Support Vector Regression}

Xu and Wang \cite{xu2012weighted} introduced weighted TSVR which gives different weights to samples and have different influence over bound functions. Computational results have demonstrated that this algorithm avoids over-fitting and also yields good generalization ability. The authors \cite{xu2014k} also proposed $K$ nearest neighbor based weighted TSVR (KNNWTSVR) which gives different penalties to the samples based on their local information on number of $K$-nearest neighbors. The weights are assigns based on number of $K$-nearest neighbors. KNNWTSVR has better accuracy but similar computational complexity as its optimization is similar to TSVR. However, the above algorithm only implements empirical risk minimization and suffer from the inverse of positive semi-definite matrices. To overcome these limitations, Tanveer et al. \cite{tanveer2016efficient} introduced an efficient regularized KNNWTSVR (RKNNWTSVR) algorithm to make it strongly convex. RKNNWTSVR leads to better generalization and robust solution. Computational cost is also reduced as it only deals with linear equations. To overcome noise sensitivity in TSVR, Ye et al. \cite{ye2016weighted} introduced weighted matrix in Lagrangian $\epsilon$ twin support vector regression. It uses quadratic loss functions and provides different weights to samples through weighted matrix. It obtains better generalization and also has less training time than TSVR and $\epsilon$-TSVR models.

\subsection{Projection Twin Support Vector Regression}

TSVR \cite{peng2010tsvr} and twin parametric insensitive SVR \cite{peng2012efficient} have obtained better generalization performance than classical SVR but both these algorithms only implement empirical risk minimization and do not include any prior information about the data samples which can lead to noise sensitivity. Thus, Peng et al. \cite{peng2014twin} proposed an efficient twin projection SVR (TPSVR) algorithm which exploits the prior structural information of data into the learning process. It seeks two projection axes such that projected points have as small as possible empirical variance values on the down-and up-bound functions. This algorithm has better generalization and requires small number of support vectors.  The aforementioned models, use uniform weighting approach and assumes that all the samples are equally important. However, this assumption may be wrong due to outliers and noise. Hence, wavelet weighted projection TWSVM for regression \cite{wang2019projection,wang2021projection} used wavelet based weights to reduce the effect of outliers.

\subsection{Robust and Sparse Twin Support Vector Regression}

Although TSVR has proven to be an effective classifier with good generalization ability, it is less robust due to the square of the 2-norm in the QPP of TSVR. In 2012, Zhong et al. \cite{zhong2012training} improved TSVR by using 1-norm rather than 2-norm distance in TSVR's QPP. It has less training time and better generalization. Chen et al. \cite{chen2012smooth} formulated smooth TSVR (STSVR) using smooth function in order to make the QPP of TSVR positive definite to obtain a unique global solution. Authors converted the QPPs to unconstrained minimization problems (UMPs) and applied Newton method to solve it effectively. All these algorithms still have high computational time due to quadratic or linear programming problems. To avoid this shortcoming, a least squares version of TSVR (TLSSVR) was formulated by Zhao et al. \cite{zhao2013twin} which leads to faster computational speed as it only deals with set of linear equations. Authors also proposed sparse TLSSVR.

In 2014, Chen et al. \cite{chen2014improved} introduced the regularization into the formulation of TSVR and implemented $l_1$-norm loss function to make it robust and sparse simultaneously. Huang et al. \cite{huang2016sparse} formulated a sparse version of least square TSVR by introducing a regularization term to make it strongly convex and also converted the primal problems to linear programming problems. This leads to a sparse solution with significantly less computational time. In 2017, Tanveer \cite{tanveer2017linear} formulated 1-norm TSVR to improve robustness and sparsity in original TSVR. 1-norm TSVR has better accuracy, generalization, and less computational time than TSVR.
In 2020, Singla et al. \cite{singla2020robust} proposed a novel TSVR (Res-TSVR) which is robust and not sensitive to noise in data. The optimization problem is non-convex with smooth $l_2$ regularizer and thus, to solve it efficiently, the authors converted it to a convex optimization problem. Res-TSVR performed best as compared to other robust TSVR algorithms in terms of robustness to noise and better generalization. Gu et al. \cite{gu2020fast} also proposed a TSVR variant suitable to handle noise called fast clustering-based weighted TSVR (FCWTSVR) which classify the samples into different categories based on their similarities and provides different weights to samples located at
different positions. The proposed algorithm performed better than TSVR, $\varepsilon$-TSVR, KNNWTSVR and WL-$\varepsilon$-TSVR. $\epsilon$-non parallel support vector regression \cite{carrasco2019epsilon}  uses two $\epsilon$-tubes for better alignment of hyperplanes and to get the more robust upbound and down bound regressor. Robust huber loss based twin SVR \cite{balasundaram2020robust} which penalizes the large deviation samples linearly and small error samples are squared. This results in robustness to noise and outliers.

\subsection{Other improvements on Twin Support Vector Regression}

TSVR has proven to provide better generalization results but it needs to solve two QPPs which increases the learning time for TSVR. Thus, Peng \cite{peng2010primal} formulated a primal TSVR (P-TSVR), which only deals with linear equations. This improves the learning speed of TSVR and shows comparable generalization. Further, to increase the sparsity of TSVR, the author introduced the back-fitting strategy for optimizing the unconstrained QPP. TSVR requires two set of constraints one with each QPP which increases the computational time for large datasets. To overcome this disadvantage, Singh et al. \cite{singh2011reduced} introduced rectangular kernels in the formulation of TSVR and proposed reduced TSVR which resulted in the significant saving of computational time and thus promoting its application for large datasets. To further reduce computational time, a LS version of TSVR (TLSSVR) was formulated by Zhao et al. \cite{zhao2013twin} which only deals with a set of linear equations. Authors also proposed sparse TLSSVR.

Huang and Ding \cite{huang2013primal} further attempted to reduce the computational time by proposing LS-TWSVM in primal space (TLSSVR) rather than dual space (PLSTSVR). This also requires to find solution of two linear equations and has comparable accuracy to TSVR. To make TSVR suitable to handle heteroscedastic noise structure, Peng \cite{peng2012efficient} proposed twin parametric insensitive SVR (TPISVR) which determines a set of nonparallel parametric-insensitive up and down-bound functions. It also works effectively when noise depends upon the input value. It requires to solve two SVM-type problems, which increases its learning speed. Computational results showed that it also has good generalization ability. Shao et al. \cite{shao2013varepsilon} implemented the SRM principle in TSVR primal space and proposed a new regressor $\epsilon$\textendash TSVR which seeks to find a pair of $\epsilon$-insensitive proximal functions. Further, to reduce complexity, the successive over-relaxation (SOR) technique is employed. Experimental results show that $\epsilon-$TSVR has better generalization and fast training speed than TSVR. 

Balasundaram and Tanveer \cite{balasundaram2013lagrangian} proposed linearly convergent Lagrangian TSVR (LTSVR) algorithm. Experimental results have exhibited its suitability and applicability due to the better generalization and less computational time than TSVR. Inspired by this algorithm, Balasundaram and Gupta \cite{balasundaram2014training} proposed Lagrangian dual of the 2-norm TSVR. Results have demonstrated an increase in learning speed with better accuracy  when compared to TSVR. Tanveer et al. \cite{tanveer2016efficient} introduced the regularization term to the objective function of TSVR and formulated regularized Lagrangian TSVR (IRLTSVR). This algorithm implements the SRM principle and requires to solve linear system of equations in place of QPP in TSVR. Optimization problems are positive definite and avoid the singularity in the solution. It has better accuracy and speed than conventional TSVR.

Balasundaram and Tanveer \cite{balasundaram2013smooth} proposed smooth Newton method for LTSVR which needs to solve linear equations in each iteration using Newton-Armijo algorithm. It has comparable generalization ability but it is at least two times faster than TSVR. Khemchandani et al. \cite{khemchandani2013twin} proposed TSVR for simultaneous learning. This algorithm is more accurate, computationally less complex and more robust as it uses $l_1$ norm error.  Lagrangian twin parametric insensitive twin SVR \cite{gupta2020lagrangian,gupta2021efficient} employed $l_2$ norm of the square variables, also it is faster as it uses linearly convergent iterative scheme for obtaining the end regressor. Asymmetric possibility and necessity regression by twin-support vector networks \cite{hao2020asymmetric} and reularization based twin SVR with huber loss \cite{gupta2021regularization} improved the generalization performance of the end regressor.

Peng et al. \cite{peng2015interval} implemented the use of interval data to handle interval input-output data (ITSVR). Rastogi et al. \cite{rastogipritam,rastogipritam1} provided an extension of $\nu$-SVR i.e $\nu$-TWSVR and large margin distribution machine based regression that it is in the true spirit of TWSVM. Balasundaram and Meena \cite{balasundaram2016training} proposed unconstrained TSVR formulation in the primal space (UP-TSVR) which is speed and obtains better generalization than TSVR. This model is solved by a gradient based iterative approach.

Parastalooi et al. \cite{parastalooi2016modified} proposed a modified version of TSVR by including the structural information from data and its distribution. Clustering is done based on the mean and covariance matrix of the data which increases accuracy. Furthermore, to increase the training speed, authors applied SOR technique and also optimized parameter selection by implementing PSO algorithm. Rastogi et al. \cite{rastogi2017nu} proposed an improved version of $\nu$-TSVR which can automatically adjust the values of upper and lower bound parameters to attain better accuracy. Experimental results have shown the superiority of the proposed algorithm over $\epsilon$-TSVR. 

TSVR gives same weights to all the samples but in fact, different positions will influence differently on the regressor which are ignored in TSVR. Thus, Xu et al. \cite{xu2018asymmetric} proposed asymmetric $\nu$-TSVR based on pinball loss function. Pinball loss function gives different penalties to the points lying above and below the bounds. It is insensitive to noise and also has better generalization ability. Tanveer and Shubham \cite{tanveer2017regularization} added regularization term in TSVR in the primal form which yields better accuracy and more robust solution.

% \begin{small}
\begin{table}[htp]
\caption {Performance of various non-linear twin support vector regression (TSVR) based algorithms. \cite{tanveer2016efficient}} \label{tab:title}
\resizebox{\textwidth}{!}{%
\begin{tabular}{p{2.5cm} p{3.2cm}p{2cm}p{2cm}p{2cm}p{2cm}}

\hline
Datasets & Models & RMSE & MAE & SSE/SST & SSR/SST \\
\hline
IBM & SVR \cite{vapnik1998statistical} & 0.1283  & 0.0920 & 0.2872 & 0.4055 \\

  & WSVR \cite{han2014weighted} & 0.0459 & 0.0379 & 0.0416 & 0.8365 \\

  & TSVR \cite{peng2010tsvr} & 0.0765 & 0.0551 & 0.1125 & 1.3937 \\
  
  & KNNWTSVR \cite{xu2014k} & 0.0330 & 0.0243 & 0.0217 & 1.0549 \\

  & RKNNWTSVR \cite{tanveer2016efficient} & 0.0330 & 0.0243 & 0.0217 & 1.0549 \\
  \hline
  
Intel  & SVR \cite{vapnik1998statistical} & 0.0500  & 0.0406 & 0.0552 & 0.8174  \\
 
  & WSVR \cite{han2014weighted} & 0.0384 & 0.0295 & 0.0330 & 1.0524 \\
  
  & TSVR \cite{peng2010tsvr} & 0.0405 & 0.0314 & 0.0368 & 0.8208\\
  
  & KNNWTSVR \cite{xu2014k} & 0.0971 & 0.0791 & 0.1922 & 0.5566\\
  
  & RKNNWTSVR \cite{tanveer2016efficient}  & 0.0382 & 0.0285 & 0.0328 & 0.8390 \\
  \hline
  
SNP-500  & SVR \cite{vapnik1998statistical} & 0.0311  & 0.0253 & 0.0192 & 0.8825  \\
 
  & WSVR \cite{han2014weighted} & 0.0296 & 0.0222 & 0.0174 & 1.064 \\
  
  & TSVR \cite{peng2010tsvr} & 0.0288 & 0.0212 & 0.0166 & 0.9702\\
  
  & KNNWTSVR \cite{xu2014k} & 0.0296 & 0.0219 & 0.0174 & 0.9904\\
  
  & RKNNWTSVR \cite{tanveer2016efficient} & 0.0273 & 0.0193 & 0.0148 & 0.9907 \\
  \hline
  
\hline
\end{tabular}}
\end{table}

% \begin{small}
\begin{table}[htp]
\caption {Performance of various non-linear twin support vector regression (TSVR) based algorithms \cite{tanveer2017regularization}} \label{tab:title}
\resizebox{\textwidth}{!}{%
\begin{tabular}{p{2.5cm} p{3.2cm}p{2cm}p{2cm}p{2cm}p{2cm}}

\hline
Datasets & Models & RMSE & MAE & SSE/SST & SSR/SST \\
\hline
Gas furnace & SVR \cite{vapnik1998statistical} & 0.0700 & 0.0459 & 0.088 & 0.8719  \\

  & TSVR \cite{peng2010tsvr} & 0.0578 & 0.0389 & 0.0599 & 0.8744 \\
 
  & LTSVR \cite{balasundaram2013lagrangian} & 0.0634 & 0.0427 & 0.0718 & 0.8689 \\
  
  & RLTSVR \cite{tanveer2017regularization} & 0.0636 & 0.0439 & 0.0718 & 0.8623 \\
  \hline
  
IBM & SVR \cite{vapnik1998statistical} & 0.1283 & 0.0920 & 0.2872 & 0.4055   \\

  & TSVR \cite{peng2010tsvr} & 0.0765 &  0.0551 &  0.1125 & 1.3937 \\
  
  & LTSVR \cite{balasundaram2013lagrangian} &  0.0328 & 0.0241 & 0.0214 & 1.0567\\
  
  & RLTSVR \cite{tanveer2017regularization} &  0.0330 & 0.0245 & 0.0217 & 1.0604 \\
  \hline
  Intel & SVR \cite{vapnik1998statistical} & 0.0500 & 0.0406 & 0.0552 & 0.8174 \\

  & TSVR \cite{peng2010tsvr} &  0.0405 & 0.0314 & 0.0368 & 0.8208 \\
  
  & LTSVR \cite{balasundaram2013lagrangian} &  0.0343 & 0.0257 & 0.0266 & 0.9180 \\
  
  & RLTSVR \cite{tanveer2017regularization} & 0.0336 & 0.0251 & 0.0255 & 0.9421 \\
  \hline
\end{tabular}}
\end{table}

\section{Applications of Twin Support Vector Regression}
\label{sec:Applications of Twin Support Vector Regression}

Ye et al. \cite{ye2013exploring} implemented $L_1-\epsilon$- TSVR for forecasting inflation. This algorithm proved to be excellent for feature ranking and determined important features for inflation in China. Experimental results showed that its accuracy is better than the ordinary least square (OLS) models. Ye et al. \cite{ye2013comparing} also implemented $\epsilon$-wavelet TSVR for inflation forecast. Authors employed the wavelet kernel that can be used for any curve in quadratic continuous integral space. This algorithm derives lower root mean squared error (RMSE) and thus, is an efficient method for inflation forecast. Ding et al. \cite{ding2013forecasting} predicted stock prices using polynomial smooth twin support vector regression. Numerical experiments reveal that this algorithm can obtain better regression performance compared with SVR and TSVR. Le et al. \cite{le2014novel} proposed a novel genetic algorithm (GA) based TSVR to improve the precision of indoor positioning. It performs better than $K$- nearest neighbor and neural network but comparable to SVR with significantly less computational time. Houssein \cite{houssein2017particle} proposed particle swarm optimization (PSO) based TSVR for forecasting wind speed. The computational results proved that it achieves better forecasting accuracy and outperformed genetic algorithm with TSVR and TSVR using radial basis kernel function. In 2018, Wang and Xu \cite{wang2018scaling} proposed safe screening rule (SSR) based on variational inequality (VI) to make TSVR efficient for large-scale problems as SSR reduces computational time significantly. Authors also implemented dual coordinate descent method (DCDM) to further increase the computational speed of TSVR. Improved twin SVR \cite{ganaiebrain2021,ganaie2021predictingbrain} was formulated for brain age prediction. Twin SVR models have been benchmarked for the prediction of brain age in Alzheimer's disease \cite{beheshti2021predicting}.

\hspace*{-2.5cm}
% \begin{small}
\begin{table}%[h]
\caption {Properties of different TSVR Algorithms} \label{tab:titleee}
\resizebox{\textwidth}{!}{%
\begin{tabular}{l c c c c }

\hline
Models $\backslash$ Characteristics & SRM & Sparsity & Matrix Inversion & Insensitive to Noise  \\\hline
 TSVR \cite{peng2010tsvr}  &  &  & \checkmark & \checkmark\\
 TWSVR \cite{khemchandani2016twsvr}  & \checkmark &  & \checkmark & \checkmark\\
 $\epsilon-$TSVR \cite{shao2013varepsilon} & \checkmark &  & \checkmark & \checkmark\\
 WTSVR \cite{xu2012weighted} &  &  & \checkmark  &  \checkmark\\
 LTSVR \cite{balasundaram2013lagrangian} &   &   & \checkmark & \checkmark\\
RLTSVR  \cite{tanveer2017regularization} &\checkmark  & & \checkmark& \checkmark\\
PTSVR  \cite{peng2010primal} &\checkmark  & \checkmark & \checkmark& \checkmark\\
KNNWTSVR \cite{xu2014k} &  & & \checkmark& \checkmark\\
RKNNWTSVR \cite{tanveer2016efficient} & \checkmark  &  & \checkmark  & \checkmark \\ 
\hline
\end{tabular}}
\end{table}

Table 3 shows the differences in major TSVR methods based on the SRM principle, Sparsity, Matrix Inversion and Noise Insensitivity.

\section{Future research and development prospects}
\label{sec:Future research and development prospects}

TWSVM classification algorithms have high training speed and accuracy than conventional SVM but it's still in the primitive stage of development and lacks practical application background. TWSVM has low generalization ability and also lacks in sparsity. Therefore, TWSVM needs further research and improvements to effectively apply to real-life challenges. Future research prospects for TWSVM  can be as follows :

\begin{itemize}
    \item An interesting aspect can be coupling other machine learning algorithms with the TWSVM. For example, a deep convolutional neural network can extract features which can be classified using TWSVM with better accuracy.

    \item There is limited research on TWSVM applications for large-scale classification. Thus, how to develop TWSVM algorithms for big data classification problems effectively is worthwhile.

     \item For non-linear classification problems, TWSVM performance highly depends upon kernel function selection and there is not enough research on this to guide researchers to choose kernels as per different applications to get desired accuracy and performance of the TWSVM algorithm. Thus, kernel selection and optimal parameters selection need further study and improvement. 

    \item Currently, only few TWSVM algorithms have been implemented for multi-class classification but it leads to class imbalance problem and often losses sparsity. Thus, further study is required for TWSVM implementation for multi-class classification.
    \item The main concept of GEPSVM/TWSVM is based on linear discriminant analysis (LDA). A well cross study on TWSVM and LDA is worthy of future work.
    
    \item
   TWSVM applications to health care is currently limited. Thus, how to implement TWSVM effectively for early diagnosis of human diseases like Alzheimer, Epilepsy, Breast cancer etc is worthy of study.
   \item
   Clustering, which aims at dividing the data samples into different clusters, is major fundamental problem in classification. Clustering based TWSVM approach is less studied currently and needs further study and development.
  \item
  There is limited research on TWSVM applications for remote-sensing. Thus, how to build efficient classifiers for remote-sensing can be explored.

    \end{itemize}
    
TSVR is much faster than conventional SVR and also has better generalization ability. But, it suffers from a lack of model complexity control and results in over-fitting. It also losses sparsity similar to TWSVM and is sensitive to outliers. Further research on TSVR can be on the following:

\begin{itemize}
\item
A significant limitation of TSVR is high computational time as it loses the sparsity. More work is required to find an efficient sparse TSVR algorithm.
\item
Data cleaning, transforming and pre-processing is an important issue for every machine learning technique and can tremendously improve results and even help in identifying novel interactions within data. As TSVR is a relatively new technique, various data handling, cleaning, pre-processing techniques like missing value imputation can be explored for improving the performance of TSVR.
\item
TSVR is evaluated only on few types of continuous variable problems. Application of TSVR can be explored on a wide range of problems in different domains.
\item
The current TSVR requires off-setting of multiple hyperparameters and hence optimal parameter selection is an issue. Thus, further research in the direction of identifying and choosing parameters should be done.

\end{itemize}

\backmatter

\bmhead{Acknowledgments}
This work is supported by Science and Engineering Research Board (SERB) funded Research Projects, Government of India under Early Career Research Award Scheme, Grant No. ECR/2017/000053 and Ramanujan Fellowship Scheme, Grant No. SB/S2/RJN-001/2016, and Council of Scientific \& Industrial Research (CSIR), New Delhi, INDIA under Extra Mural Research (EMR) Scheme grant no. 22(0751)/17/EMR-II.  We gratefully acknowledge the Indian Institute of Technology Indore for providing facilities and support. Yuan-Hai Shao acknowledges the financial support by the National Natural Science Foundation of China (11871183, 61866010, 11926349).

\bibliography{refs}% common bib file
%% if required, the content of .bbl file can be included here once bbl is generated
%%\input sn-article.bbl

%% Default %%
%%\input sn-sample-bib.tex%

\end{document}